\documentclass[11pt]{article}

\usepackage{xcolor,fullpage}
\usepackage{ulem}
\usepackage{pslatex}
\usepackage{graphicx}
\usepackage{fancyhdr}
\usepackage{amsmath}
\usepackage{amssymb}
\usepackage{textcomp}
\usepackage[numbers,sort&compress]{natbib}	
\usepackage{subcaption}	
\usepackage{hyperref}
\usepackage{setspace}
\usepackage{cancel}
\usepackage{mathtools}
\usepackage[font=footnotesize]{caption}
\usepackage{subcaption}
\usepackage{lscape}
\usepackage{rotating}
\usepackage{adjustbox}
\usepackage{multirow}
\usepackage{tabularx}
\usepackage{lipsum}
\usepackage{booktabs}
\usepackage{subfiles}
\usepackage{titling}
\usepackage{placeins}
\usepackage{authblk}
\usepackage{comment}
\usepackage{wrapfig}

\newcolumntype{L}[1]{>{\raggedright\let\newline\\\arraybackslash\hspace{0pt}}m{#1}}
\newcolumntype{C}[1]{>{\centering\let\newline\\\arraybackslash\hspace{0pt}}m{#1}}
\newcolumntype{R}[1]{>{\raggedleft\let\newline\\\arraybackslash\hspace{0pt}}m{#1}}

\title{Learning Latent Dynamics via Invariant Decomposition and (Spatio-)Temporal Transformers}

\author[1]{Kai Lagemann$^*$}
\author[2]{Christian Lagemann$^*$}
\author[1,3]{Sach Mukherjee}
\affil[1]{\small Statistics and Machine Learning, DZNE, Bonn, Germany}
\affil[2]{Institute of Aerodynamics, RWTH Aachen University,  Aachen, Germany}
\affil[3]{MRC Biostatistics Unit, University of Cambridge, Cambridge, UK}
\affil[$*$]{{\footnotesize  Joint first authors}}

\begin{document}
\date{}
\maketitle

\vspace{-2cm}

\begin{abstract}
\normalsize
\noindent
We propose a method for learning dynamical systems from high-dimensional empirical data that combines variational autoencoders and (spatio-)temporal attention
within a  framework designed to enforce certain scientifically-motivated invariances.
We focus on the setting in which data are available from multiple different instances of 
a system whose underlying dynamical model is entirely unknown at the outset.
The approach rests on a separation into an instance-specific encoding (capturing initial conditions, constants etc.) and a latent dynamics model
that is itself universal across all instances/realizations of the system.
The separation is achieved in an automated, data-driven manner and only empirical data are required as inputs to the model.
The approach  allows effective inference of system 
behaviour at any continuous time but does not require an explicit neural ODE formulation, which makes it efficient and highly scalable.
We study  behaviour through simple theoretical analyses and  extensive experiments on synthetic and real-world datasets. The latter investigate  learning the dynamics of  complex systems based on finite data and show that the proposed approach can outperform state-of-the-art neural-dynamical models. 
We study also more general inductive bias in the context of transfer to data obtained under entirely novel system interventions. 
Overall, our results provide a promising new framework for efficiently learning dynamical models from heterogeneous data with potential applications in a wide range of fields including physics, medicine, biology and engineering.

\end{abstract}

\section{Introduction}
\begin{figure}[tbh]
\centering
\includegraphics[width=\textwidth]{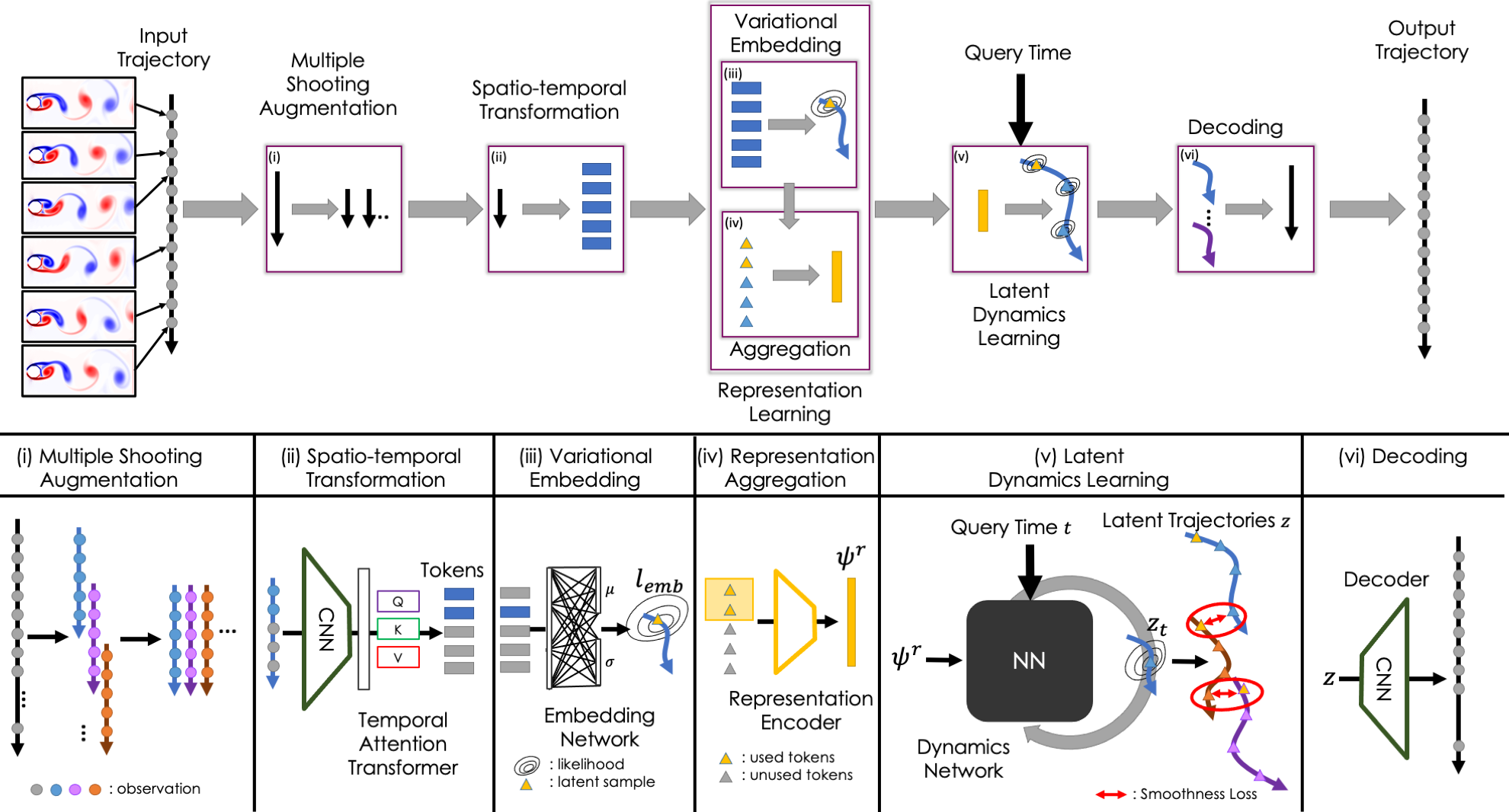}
\caption{Architecture: Learning of Latent Dynamics via Invariant Decomposition (LaDID). Top half: general flow chart of the compute steps (i)-(v) of LaDID; lower half: more detailed representation of the compute steps. A set of high-dimensional snapshots of a system on a regular or irregular time-grid serves as the empirical input to LaDID. The trajectory is split into subpatches using Multiple Shooting Augmentation (i). The first time-points of each subpatch are used to compute a subtrajectory representation: features of the selected snapshots are re-weighted wrt time and spatial location (ii), transformed to low-dimensional variational embedding (iii), and aggregated into one trajectory representation $\psi_r$ (iv). During inference, the latent dynamical model is conditioned on the specific representation $\psi_r$. Prediction is possible at any continuous time by querying the latent state of any time point of interest (v). Latent subtrajectories are sewn together by a smoothness loss. Finally, the entire latent trajectory is decoded to the observation space (vi).}

\end{figure}

Dynamical models are central to our ability to understand and predict natural systems.
For many systems of interest in fields like biology, medicine and engineering, high-dimensional observations can reasonably be thought of as obtained via dynamics operating in a lower dimensional space and this assumption (related to the manifold hypothesis \cite{fefferman2013ManifoldHypothesis}) is a common one in many settings.
Machine learning (ML) approaches for learning dynamical systems have been an important area of recent research, including in particular neural ordinary differential equations \cite{chen2018NODE} 
and a wider class of related neural-dynamical  models (\cite{zhi2022learning,finlay2020train,duong2021hamiltonian,choi2022learning,chen2020learning,kim2021inferring}). These models define layers as differential equations and in that sense incorporate an  informative (and often  appropriate) inductive bias for physical systems.

In this paper, we propose a new framework for learning latent dynamics from observed data. 
Our approach, called ``Latent Dynamics via Invariant Decomposition'' (LaDID), combines variational autoencoders and spatio-temporal attention
within a learning framework motivated by certain scientifically-motivated invariances.
The LaDID framework exploits these invariances 
and provides model output at any continuous time
but does not require an explicit  ODE formulation (details below).
Although the focus of this paper is primarily on methodology, the research we present is motivated by real-world applications, particularly in the fields of biomedicine and health. In these domains, it is often the case that explicit dynamical models are not initially accessible; however we expect that 
the unknown dynamical models 
 should nevertheless satisfy the very general invariances outlined below (see also Discussion).

The methods we propose build on two observations concerning classical scientific models:
\begin{itemize}
\item First, the notion that every output from a class of scientific systems should be explainable via a single model despite differences between individual instances within a class. In this sense, the model is considered universal, even if there are large differences in observed data across instances. For instance, in the case of mechanical systems, one set of equations can effectively describe diverse instances (e.g. by accounting for dissimilar initial conditions or constants). This is  intriguing from an ML perspective since the distribution of data can vary greatly between instances, leading to significant shifts.
\item Second, instance/realization-specific factors (such as 
initial conditions or constants) tend to remain unchanged throughout the entire duration of a given realization. In this regard, these factors exhibit time invariance, as they maintain their values consistently over time.
\end{itemize}

Our approach and the derived transformer-based architecture builds 
directly on these notions of invariance. 
We describe the network in detail below, but in brief the set-up is as follows:
  From an available trajectory -- thought of as representing a specific instance/realization $r$ of a more general model class $\mathcal{M}$ -- we learn an encoding $\psi^r$ of the realization-specific information.
This is intended to 
implicitly 
capture information (such as initial conditions or constants)
that are specific to the instance or realization $r$, but the information should remain valid for all times within a realization/instance; hence the encoding has a superscript indicating the realization but 
no time index.
This encoding $\psi^r$ is treated as an input to a ``universal'' model $f$ to enable prediction of system output at any time $t$. The  model $f$ itself is learned across multiple realizations $r$ of the system defined by the model class but the same function is always used for prediction (for 
the system $\mathcal{M}$).
In other words, $f$ is intended to be universal (across all queries concerning the system $\mathcal{M}$)
with realization-specific information provided only by the input $\psi^r$.
We argue that under certain conditions, this decomposition into {\it realisation-specific} (RS)
and {\it realization-invariant} (RI) information 
allows for 
definition of a simple and 
convenient learning framework. We propose a deep neural architecture for this purpose, showing how learning of 
universal latent dynamics and realization-specific information can be done jointly, in an end-to-end manner.
This enables prediction  of system behaviour at {\it any} continuous time $t$, for {\it any} realization $r$ (for which minimal data are available).

We empirically validate our proposed method on spatio-temporal systems with dynamics governed by ordinary or partial differential equations. These systems are ubiquitous in nature and include physical phenomena in rigid body motion, fluid dynamics and turbulent flows, electromagnetism and molecular dynamics.
Our work is related to a large body of previous work on neural learning of dynamical models, which we discuss in detail below. A key distinction of our approach is that by leveraging the framework outlined above, we do not require 
an explicit neural ODE (NODE) at all; rather we can carry out learning within a simple, broadly supervised framework that, as we show, can in fact outperform existing neural-dynamical models   over a range of challenging tasks on both regular and irregular time grids.

Thus, the main contributions of this paper are:
\begin{itemize}
    \item We present a novel framework, and associated transformer-based network,   for the 
    separation and 
    learning of realization-specific information and (realization-invariant) latent dynamical systems. 
    \item We systematically study performance on short- and longer-horizon prediction of a wide range of complex temporal and spatio-temporal problems, comparing against a range of state-of-the-art neural-dynamical methods. 
 \item We study the challenging case of transfer to data obtained under entirely novel system interventions via a few-shot learning  approach. 
\end{itemize}

The remainder of the paper is organized as follows. We first introduce the problem statement, basic conceptual ideas and assumptions and the proposed  architecture. 
We then present  empirical work, spanning a wide range of different, challenging dynamical systems.
We close with comments on limitations and future directions.

\section{Related Work}
We begin with a short survey of
the rich body of recent work on  ML-based approaches  aimed at  learning dynamical models from data. \cite{brunton2016DiscoveringGoverningEquationsFromData, kaheman2020SINDyPi} access Koopman theory to learn coefficients of a finite set of non-linear functions to approximate the observed dynamics. Other approaches \cite{cranmer2020LagrangianNeuralNetworks, lutter2019DeepLagrangianNets, greydanus2019HamiltonianNeuralNetworks, finzi2020SimplifyingHNNLNNwithConstraints, zhong2020SympleticODENet-HNNwithControl, bai2019DeepEquilibriumModels} induce Hamiltonian and Lagrangian priors into neural networks exploiting the reformulations of Newton’s equations of motion in energy-conservative dynamics. This  inductive bias allows for accurate predictions over a significantly wider time-horizon but involves strong assumptions.
Purely data-driven ML approaches have attracted much recent attention and have shown strong performance in many areas. \cite{heinonen2018ODEwithGP} learn a data-driven approximation of a dynamical system via a Gaussian Processes and \cite{li2020FourierNeuralOperator} formulates a new neural operator by parameterizing
the integral kernel directly in Fourier space to allow zero-shot prediction for an entire family of PDEs. 

In a recent line of research, the 
ability of NNs to approximate arbitrary functions
has been exploited for unknown dynamical models, with connections drawn between deep architectures and numerical solvers for ODEs, PDEs and SDEs \cite{chen2018NODE, Weinan2017MLviaDynamicalSystems, lu2020BeyondFiniteLayerNN, ruthotto2018NNbyPDE, Haber2017StablArchsForNN, richterpowell2022neuralConservationLaws}. This has given rise to new algorithms rooted in neural differential equations that have been shown to provide improved modeling of a variety of dynamical systems relative to standard recurrent
neural networks (RNNs) and their variants. Improvements for neural ODEs have been proposed: \cite{dupont2019AugmentedNODE} put forward a simple augmentation to the state space to broaden the range of dynamical systems that can be learned; \cite{kidger2020CDEforIrregularTime} propose neural controlled differential equations; 
\cite{morill2020NeuralRoughODEForLongTime} incorporate concepts of rough theory to extend applicability to trajectories spanning thousands of steps, and \cite{jia2019NeuralJumpSDE} explore extensions to account for interrupting stochastic events. The work of \cite{rubanova2019ODERNN}  combines an autoregressive RNN updated at irregularly sampled time points with an NODE model that computes inter-observation states in continuous time. Their work also includes a variational autoencoder (VAE) extension to uncover latent processes based on high-dimensional observations. In related work, \cite{yıldız2019ode2vae} proposed a VAE for sequential data with a latent space governed by a continuous-time probabilistic ODE. \cite{massaroli2021differentiableMultipleShootingLayers} transferred the concept of multiple shooting to solve differential equations to the conceptual space of NODEs and \cite{iakovlev2022MSVI} extended this concept to sparse Bayesian multiple shooting, with both works evaluating latent NODEs.

Our approach is inspired by this body of work in that we also use neural networks to learn latent  dynamical models. However, two key differences are as follows. First, our models are specifically designed  to exploit certain invariances that are important in classical scientific models (as detailed below). From this point of view, we leverage a particular kind of scientifically-motivated inductive bias from the outset. Second, exploiting these invariances allows us to eschew explicit  neural ODEs altogether, providing an arguably simpler, transformer-based scheme that 
can be trained in a straightforward fashion, but that, as we show,
is effective across a range of dynamical problems and that 
 can even be leveraged for few-shot learning to generalize to 
nontrivial system interventions.

\section{Methods} \label{sec: Method}
We first provide a high-level problem statement.
We then consider some simple arguments 
to support the idea that a relatively straightforward learning framework can be applied in 
this setting. 
These simple but general arguments are intended to 
shed light on the 
learning problem and 
provide conceptual background. 
We then put forward a specific architecture to permit learning in practice, i.e. a concrete learning framework that we implement and study empirically. 

\subsection{Problem statement}

We focus on settings in which available training data corresponds to multiple instances or realizations 
of a system class of interest. We seek to learn a model that allows prediction of the system at {\it any} continuous time  for {\it any} realization/instance  (this set-up is common in many areas of science, engineering and medicine, and we study several specific examples in results below).
Let $X_{r,j,t} \! \in \! \mathbb{R}^p$ denote observed data for realization $r$ at time $t$, with $j$ indexing the dimension.
In the training phase, we assume that for each of $R$ different realizations we have access to an observed trajectory of $T$ time points. We need to know the sampling times themselves, but these need not be evenly spaced.
Hence, the input $X$ is of size $R {\times} p {\times} T$.
We treat the $p$-dimensional observations as arising from an underlying process $f^*$ operating in $q {<} p$ dimensions coupled with a time-independent observation process $g: \mathbb{R}^q \rightarrow \mathbb{R}^p$. Our goal is to learn a latent dynamical model $f$ that allows prediction of $X_{r,j,t}$ at any time $t$ and for any realization $r$. For realizations that were not represented in the training data at all, we assume availability of some data specific to the realization at inference-time. 

\subsection{From invariances to a simple learning framework}
In this Section, we consider an abstract version of the problem of interest, aimed at clarifying the specific invariances that will be needed and gaining understanding of how learning can be conveniently performed in this setting. 
The notation in this Section is self-contained but
(in the interests of expositional clarity)
 differs from the following Sections providing architectural and implementation details.

{\it General formulation}.
Consider an entirely general system $f$ which may be deterministic or stochastic (with all random components absorbed for convenience into $f$). 
We are interested in settings in which some aspects of the model are realisation-specific (RS) while others remain realisation-invariant (RI). Let $$x^r_t {=} f(t;\Theta_r), \, x^r_t {\in} \mathbb{R}^p$$ denote the fully general model. 
Here, $\Theta_r$ is the complete parameter  needed to specify the time-evolution, including both RS and RI parts. 
To make the separation clear, we write the two parts separately as $x^{r}_t {=} f(t;\theta_r, \theta)$, where $\theta_r$, $\theta$ are respectively the RS and RI parameters (together comprising $\Theta_r$).

We call this a {\it generalized initial condition formulation}, as it generalizes the idea of an initial condition in ODEs. In the case of an ODE, the initial conditions and relevant constants are the information needed, in addition to the model equations themselves, to fully specify the time evolution of any specific instance/realization of the model. In our terms, if a problem/dataset has one model but many such ``initial conditions'' (more precisely this can be any RS aspect, including constants), then the model itself is RI, while the initial conditions/constants are the RS part. Note that 
although we do not assume any specific knowledge about the system (other than the motivating invariances),
we do assume that we
can at the outset block datasets into instances arising from a shared system (whose details are entirely unknown); in this paper we do not consider the task of learning the system classification itself from data.

{\it Fully observed case}.
In the model $f$ above, the true parameter $\Theta_r = (\theta_r, \theta)$ comprises RS and RI parts.
We now provide conditions under which learning of the system state at any continuous time $t$ is possible
without explicit knowledge of either the model $f$ or the true parameter 
$\Theta_r$.
We start with the simplest case of fully observed data (i.e. no latent dynamics) and then consider the latent case.
The idea is to work from a candidate encoding $\hat{\theta}_r$ of the RS information. In practice this would be the output of a neural network (NN) based on initial data from a realization $r$ and itself learned end-to-end jointly with other model components; see subsequent Sections for 
architectural and implementation details. The encoding $\hat{\theta}_r$ is intended to be a representation of RS information that, as we will see below, under certain assumptions can be combined with a universal model to allow effective prediction.
 
Specifically, we assume that the encoding $\hat{\theta}_r$, while possibly incorrect (i.e. such that $\hat{\theta}_r {\neq} \theta_r$) satisfies the property
$$\exists m, \exists \theta_m : \theta_r {=} m(\hat{\theta}_r;\theta, \theta_m),$$
where $m$ is a function that ``corrects'' $\hat{\theta}_r$ to give the correct RS parameter.
Note that the correction function $m$ can potentially use the RI parameter of the system and possibly additional parameters $\theta_m$. This essentially demands that while the encoding $\hat{\theta}_r$ might be very different from the true value (and may even diverge from it in a complicated way that depends on unknown system parameters), there exists an RI transformation that recovers the true RS parameter from it, and in this sense the encoding contains all RS information. 
We call this the \textit{sufficient encoding assumption} (SEA). Note that the function $m$ has an oracle-like
property in that it may depend on the true RI parameter $\theta$ and we will not have access to $m$ in practice.

We would like to learn a mapping that takes as input the RS encoding 
$\hat{\theta}_r$ and query time point
$t$ and yields the correct system output (for any 
realization and any time).
That is, the input to the mapping would be the pair $(t,\hat{\theta}_r)$ and the output
intended to predict $x_t^{r}$.
To this end, consider the candidate prediction
$$\hat{x}_t^{r}  {=}  f(t ; m(\hat{\theta}_r; \theta,\theta_m), \theta),$$ 
and observe that we can always write the RHS as $h(t,\hat{\theta}_r; \Xi)$ 
where $\Xi = (\theta, \theta_m)$ is a RI parameter
and $h$ is a function (obtained by combining $f$ and $m$ as above).
This latter formulation emphasizes the fact that the RHS is in fact a function (here, $h$) of only the inputs $(t,\hat{\theta}_r)$, and therefore potentially learnable from training pairs. Note that the parameters of $h$ are entirely RI and hence the only RS information is carried by the encoding $\hat{\theta}_r$. It is easy to see that under SEA this construction provides the correct output since we can write the RHS as 
$f(t ; m(\hat{\theta}_r; \theta,\theta_m), \theta) = f(t; \theta_r, \theta) = x_t^{r}$.
Thus, combining encoding $\hat{\theta}_r$ and function $h$ allows prediction of the time evolution of any realisation.  In other words, even if the RS encoding is distant from the true RS parameter, 
under SEA there exists a RI function that can correct it, and 
we can therefore seek to train a NN
aimed at learning a function $h$ which combines these RI elements to provide the desired mapping.

\medskip 

{\it Latent dynamics}.
In line with the manifold hypothesis, consider now dynamics at the level of latent variables $z \in \mathbb{R}^q$ and again consider a model with RS and RI parts but at the level of the latents, i.e. $z^r_t = f(t, \hat{\theta}_r; \theta_r, \theta)$. We assume that the observables are given by (an unknown) function of the hidden state $z$. We first consider the case in which the latent-to-observed mapping is RI, and then the more general case of a RS mapping.

{\it Case I: RI mapping.} 
Assume the observable is given as $x^r_t = g(z^r_t; \theta_g)$, where $g : \mathbb{R}^q \rightarrow \mathbb{R}^p$ is the (true) observation process and $\theta_g$ is an RI parameter. Further, assume that we have an estimate $\hat{\theta}_r$ of the RS encoding that satisfies the sufficient encoding assumption (SEA). In a similar fashion, assume we have an estimate $\hat{\theta}_g$, which may be incorrect (in the sense of $\hat{\theta}_g \neq \theta_g$) but that satisfies:
$\exists m_g, \exists \theta_{m_g} : \theta_g = m_g(\hat{\theta}_g;\theta, \theta_{m_g})$.
That is, $\hat{\theta}_g$ admits an RI correction. As above, the correction is oracle-like and may potentially depend on true RS parameters. Note also that subject to the existence of a correction the estimate (and implied mapping) may be potentially arbitrarily incorrect. In analogy to SEA, we call this the \textit{sufficient
mapping assumption} (SMA).

Now, consider  training of a NN, with training (input, output) pairs of the form
$\{(t, \hat{\theta}_r), x_t^{r}   \}_{(t,r)\in Train}$.
We want to understand whether supervised learning of a universal model to predict output for arbitrary queries $(t, r)$ is possible. This is not obvious, since we now have training data only at the level of the observables, but the actual dynamics operate at the level of latents. 
Consider the following function $h_{SMA}$:
\begin{equation*}
    h_{SMA}(t, \hat{\theta}_r; \Xi) = g(f(t; m(\hat{\theta}_r; \theta, \theta_m);m_g(\hat{\theta}_g; \theta, \theta_{m_g}))
\end{equation*}
where $\Xi = (\theta, \theta_m, \theta_{m_g})$ is an RI parameter.

Under SEA and SMA it is easy to see that $h_{SMA}$ provides the correct output, since:
\begin{align*}
    h_{SMA}(t, \hat{\theta}_r; \Xi) &= g(f(t; m(\hat{\theta}_r; \theta, \theta_m);m_g(\hat{\theta}_g; \theta, \theta_{m_g}))\\
    &= g(f(t; \theta_r, \theta); \theta_g)\\
    &= g(z^{r}_t; \theta_g)\\
    &= x^{r}_t
\end{align*}
That is, under SEA and SMA there exists a 
function 
of $t$ and $\hat{\theta}_r$ that provides the correct output
and that is universal in the sense that (i) the same function applies to any query $(t,r)$ and 
(ii) its parameter is itself 
RI and hence the same for all realizations.

{\it Case II: RS mapping.} Suppose now the mapping is RS, with the model specification as above but with the observation step $x^{r}_t = g(z^{r}_t; \theta^{r}_g)$, where $\theta^{r}_g$ is an RS parameter. This means that the
latent-observable relationship is itself non-constant and instead varies between realizations.

Assume we have a candidate estimate $\hat{\theta}^{r}_g $ which may
be incorrect in the sense of $\hat{\theta}^{r}_g \neq \theta^{r}_g$ but that satisfies: 
$\exists m_g, \exists \theta_{m_g} : \theta_g = m_g(\hat{\theta}^{r}_g;\theta, \theta_{m_g})$.
 In analogy to SMA, we call this the {\it realization-specific sufficient mapping assumption} or RS-SMA. Now to create training sets, we extend the formulation to require input triples, as:
$\{(t, \hat{\theta}_r, \hat{\theta}_g^{r}), x_t^{r}   \}_{(t,r)\in Train}$.
In a similar spirit to the RS encoding above, 
$\hat{\theta}_g^{r}$ in the input triples may be incorrect, but only needs to satisfy RS-SMA. As in Case I, we have training data only at the level of observables (not latents) but want to understand whether supervised learning of a model to predict output for arbitrary queries $(t, r)$ is possible.
Consider the function $h_{RS-SMA}$:
\begin{equation*}
    h_{RS-SMA}(t, \hat{\theta}_r,\hat{\theta}_g^{r}; \Xi) = g(f(t; m(
\hat{\theta}_r; \theta, \theta_m), \theta); m_g(\hat{\theta}_g^{r}; \theta, \theta_{m_g}))
\end{equation*}
where $\Xi = (\theta, \theta_m, \theta_{m_g})$ is an RI parameter.
In a similar manner to Case I, it is easy to see that $h_{RS-SMA}$ provides the correct output under SEA and RS-SMA, since:
\begin{align*}
    h_{RS-SMA}(t, \hat{\theta}_r,\hat{\theta}_g^{r}; \Xi) &= g(f(t; m(\hat{\theta}_r; \theta, \theta_m), \theta); m_g(\hat{\theta}_g^{r}; \theta, \theta_{m_g}))\\
 &= g(f(t; \theta_r, \theta); \hat{\theta}_g^{r})\\
 &= g(z^{r}_t; \hat{\theta}_g^{r})\\
 &= x^{r}_t
\end{align*}

\medskip

The foregoing  arguments are based on 
an abstract view of the task at hand
and show that under the assumptions above, 
there exist universal mappings from the available inputs to the desired outputs whose parameters are themselves RI.
As a result, subject to the assumptions above, 
it may be possible to learn suitable mapping functions from data (without requiring prior access to the various components). We now put forward a specific architecture   aimed at learning 
such a mapping in practice.

\subsection{Neural architecture}

In this Section we provide details on the NN architecture and implementation.
Additional notation is required to clarify the relevant details;
the exposition below is self-contained and intended to be concise, however  the notation differs from that in Section 3.2 above.
We note also that the architecture introduced in this Section is only one approach by which to instantiate the general ideas above.

We seek to learn a model able to predict future observations $x_{t:T}^r$ based on a sequence of $k$ past time points $x_{t-k:t-1}^r$. 
 To this end, we use a variational autoencoder (VAE)  \cite{kingma2013autoencoding, rezende2014stochastic} within our model. 
From a high-level point of view, it consists of an encoder network that learns to embed a set of observations of 
different realisations $X_{1:T}^r$ to a low-dimensional embedding $z_{1:T}^r \sim q_{\Phi_L}(\cdot | X_{1:T}^r)$ 
and $\Phi_L$  being the parameter set of the transformation to the latent space.
The dynamics are modelled in this inferred state space stochastically. Finally, a decoder network maps the position coordinates in the state space back to the image/observation space $p(X_{1:T}^r) = \mathcal{D}_{\phi_{dec}}(z_{1:T}^r)$. 

In more details, the encoder is a collection of three NNs. First, features from the input observations $x_{t-k:t-1}^r$ are extracted using a {\it convolutional} neural network (CNN) parameterized by $\phi_{enc}$ shared across all representations and patches  (see {\it multiple shooting augmentation} below). Our CNN encoder transforms the sequence of input observations to a sequence of feature vectors, $z^{(enc),r}_{t-k:T}{=}f_{\phi_{enc}}(x^r_{t-k:T})$. Then, we compute the trajectory representation and the latent embedding as follows. Each input patch is split into two disjoint sets by time. The first $k$ data points $\mathcal{M}_R{=}\{z^{(enc), r}_{t-k:t}\}$ are used to compute a trajectory specific representation distribution $\psi^r {\sim} q_ {\Phi_R}(x_{t-k:t-1}^r) = \mathcal{N}(\mu_r, \sigma_r)$ and $\mu_r, \sigma_r {=} f_{\Phi_R}(z^{(enc), r}_{t-k:t})$. In cases of irregularly sampled trajectories, we use a time threshold $\tau$ to define the representation set, $\mathcal{M}_R {=} \{z^{(enc), r}_{t_i}\}, \, t_i {\in} \{t{<}\tau\} $.  We model $f_{\phi_R}$ as a transformer network with temporal attention. In other words, we consider the sequence feature vectors $z^{(enc), r}_{t-k:t}$ as an time-ordered sequence of tokens and transform each token according to the temporal distance to the other tokens. The trajectory representation $\psi^r$ is the {\it mean}-aggregation of the temporally transformed representation tokens. 
With initial density given by the encoder networks $q_{\Phi_L}(z_t|x^r_{t-k:T}, \psi^r)$, the density for all queried latent points (on a continuous time grid) can be
predicted by $z_{t_q}^r {\sim} \mathcal{N}(\mu_{t_q}^r, \sigma_{t_q}^r)$ with $\mu_{t_q}^r, \sigma_{t_q}^r {=} f_{\phi_{dyn}}(t_q,z_t,\psi^r)$. Note that this approach allows for latent state predictions at any time since the learned dynamics module $f_{\phi_{dyn}}$ is continuous in time and our variational model utilizes encoders only for obtaining the initial latent distribution.  We also make use of the reparameterization trick to tackle uncertanties in both the latent states and in the trajectory representations \cite{kingma2013autoencoding}. Finally, the decoder maps the latent state to the output. In summary, our variational model is defined as follows:

\begin{table}[h!]
\centering
\begin{adjustbox}{max width=\textwidth}
\begin{tabular}{ll|ll}
\multirow{2}{*}{Encoder}   & $\mu_r, \sigma_r=f_{\phi_R}((f_{\phi{enc}}(x^r_{t-k:t})$                     & \multirow{2}{*}{Reparameterization} & $\psi^r \sim q_{\Phi_R}(\psi^r|x^r_{t-k:t})$                                      \\
                           & $\mu_t, \sigma_t = f_{\phi_{dyn}}(f_{\phi_{enc}}(x^r_{t})$                    &                                     & $z_t \sim q_{\Phi_L}(z_t|x^r_{t},\psi^r)$                                     \\ 
\hline
\multirow{2}{*}{Posterior} & $q_{\Phi_R}(\psi^r|x^r_{t-k:t})= \mathcal{N}(\mu_r, \sigma_r)$    & \multirow{2}{*}{Decoder}            & \multirow{2}{*}{$\hat{x}_{t_q}^r = f_{\phi_{dec}}(f_{\phi_{dyn}}(t_q, z_t, \psi^r) ~ t_q>t$}  \\
                           & $q_{\Phi_L}(z_t|x^r_{t},\psi^r)=\mathcal{N}(\mu_t, \sigma_t)$ &                                     &                                                                                                  
\end{tabular}
\end{adjustbox}
\end{table}

\paragraph{Multiple shooting augmentation.} To effectively process longer-horizon time series data, we apply a variant of {\it multiple shooting}. However, since our model does not rely on explicit ODE formulation, we are not concerned with turning an initial value problem into a boundary value problem \cite{massaroli2021differentiableMultipleShootingLayers}. Instead, our method decomposes each realisation  $x_{t:T}^r$ into a set of $N$ overlapping subtrajectories, independently reduces each patch to a latent representation, and learns to model the dynamics on each patch using neural networks. The patches are sewn together by a Bayesian smoothness prior to obtain a global trajectory. We set the time of the start point of each patch to $t_{n, 1:T_n}=0$ and recompute the timings of the remaining observations relatively wrt $t_{n, 1}$.  We define the prior for our model as 
\begin{align}
    p(z_{1:N}, \psi_{1:N})=p(z_{1:N}|\psi_{1:N})p(\psi_{1:N})
\end{align}
where $p(\psi_{1:N})$ are Gaussians, and $p(z_{1:N}|\psi_{1:N})$ denotes the smoothness prior defined as $p(z_{1:N}|\psi_{1:N}) = p(z_1|\psi_1)p(\psi_1)\prod_{n=2}^N p(z_n | z_{n-1}, \psi_n) p(\psi_n)$. Intuitively, this prior stitches the first point of each patch to the last point of the previous patch given the corresponding representation. 

\paragraph{Optimization.} We optimize our model by variational inference meaning that we 
maximize the evidence lower bound (ELBO):
\begin{align}
    \max_{\Phi} ~ &\underbrace{\mathbb{E}_{\psi^r\sim q_{\Phi_R}, z_t\sim q_{\Phi_L}}\sum_{n=1}^N \text{log}~ p_n(\hat{x}_n)}_\text{(i) likelihood term} - \underbrace{\sum_{n=1}^N\text{KL}(q_{\Phi_R}\| \mathcal{N}(0,1))}_\text{(ii) representation prior}\nonumber\\ 
    &- \underbrace{\sum_{n=2}^N \mathbb{E}_{\psi^r\sim q_{\Phi_R}, z_t\sim q_{z_{n-1}, \psi^n}}\left[\text{KL}[q(z_n)\|p(z_n | z_{n-1}, \psi_n)]\right]}_\text{(iii) smoothness prior},
    \label{Eq.: Loss}
\end{align}
with $\Phi_R$ and $\Phi_L$ denoting collections of corresponding parameters for computing representations and latent dynamics. The expectations are approximated by single-sample Monte Carlo integration. 

\section{Datasets}\label{Sec: Datasets}
To study the capabilities of LaDID relative to existing models, we consider several physical systems ranging from relatively simple ODE-based datasets to complex turbulence driven fluid flows. Specifically, we evaluate LaDID  on high-dimensional observations ($p{=}16,384$) of a nonlinear swinging pendulum, the chaotic motion of a swinging double pendulum, and realistic simulations of the two-dimensional wave equation, a lambda-omega reaction-diffusion system, the two-dimensional incompressible Navier-Stokes equations, and the fluid flow around a blunt body solved via the latticed Boltzmann equations. This extensive range of applications covers datasets which are frequently used in literature for dynamical modeling and therefore enable fair comparisons to state-of-the-art baselines and at the same time study effectiveness in the context of complex datasets relevant to real-world use-cases. To this end, we evaluate LaDID on regular and irregular time grids and further transfer a learned model prior to a completely unknown setting obtained by {\it intervention} on the system. That is, we generate small datasets on intervened dynamical systems (either via modifying the underlying systems, for example by changing the gravitational constant or the mass of a pendulum, or via augmenting the 
realisation-specific observation, e.g. by changing the length of the pendulum or the location of the simulated cylinder) and fine-tune a pre-trained model on a fraction of the initially seen training data. Please refer to Appendix \ref{Appendix: Dataset details} for further details on the datasets. 

\section{Experimental Setup} \label{Sec: Setup}
\begin{figure}[tb]
\centering
\includegraphics[width=\textwidth]{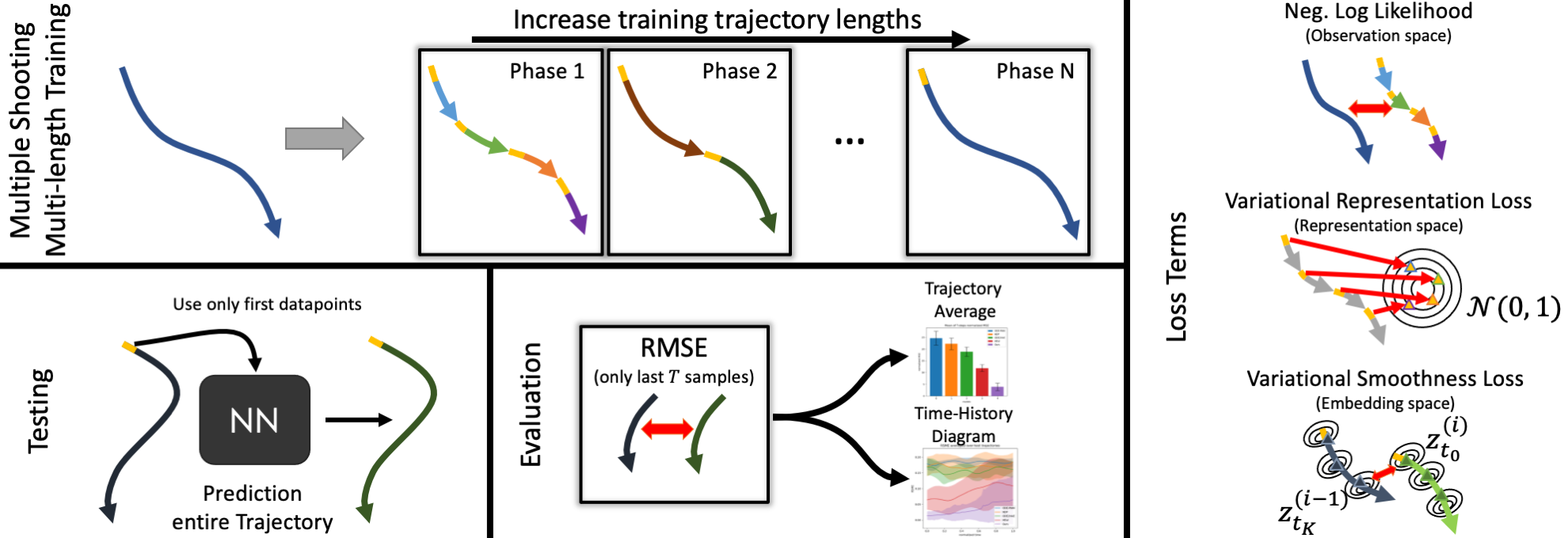}
\caption{Training scheme with losses, and test/evaluation procedure. Top left: Multiple Shooting Multi-length Training. An input trajectory is split into subpatches. Subtrajectory length is increased in multiple phases to the length of the input trajectory. 
Bottom left: Testing: only the first few points are used to roll-out the latent trajectory and transformed to the observational space. Evaluation: Last $T$ samples of the predicted trajectory are used to compute the evaluation metrics, the average of the summed normalized mean squared error and a time history diagram showing the error evolution.
Right: Loss consisting of three parts: negative log likelihood loss to penalize reconstruction errors, representation loss to define a gradient field between representations, smoothness loss to penalize jumps between latent subpatches. } \label{Fig: Training Testing Losses}
\end{figure}
\paragraph{Training.} 
We train all experiments in a multi-phase schedule wrt the multiple shooting loss in eq. \ref{Eq.: Loss}. In the different phases, we split the input trajectory into overlapping patches and start learning by predicting one step ahead. We double the number of prediction steps per patch every 3000 epochs meaning that learning is done on longer patches with decreased number of patches per trajectory. In cases where the trajectory length is not divisible by the number of prediction steps, we drop the last patch and scale the loss accordingly. In the final phase, training is carried out on the entire trajectory. All network architectures are implemented in the open source framework PyTorch \cite{paszke2019pytorch}. 
Further training details and hyperparameters can be found in Appendix \ref{Appendix: Training details}.

\textbf{Testing.}   We test the trained models on entirely unseen trajectories. During testing, we only provide the first $k{=}10$ trajectory points to the learned model. Based on these samples, we compute a trajectory representation $\psi^r$ followed by rolling out the latent trajectory at the time points of interest. Finally, we compare  predictions and ground truth observations by evaluation metrics.

\textbf{Evaluation metrics.} \label{Subsec: Metric}   Our experiments focus on two key metrics. Firstly, we calculate the mean squared error (MSE) for extrapolated trajectories, where we evaluate the model's performance over a total of $2T$ steps and measure the MSE over the last $T$ timesteps. This approach allows us to assess whether extrapolation over future time periods can better predict the model's ability to extrapolate further in time, compared to reconstruction MSE. The value of $T$ is set to 60 for all our experiments, and we normalize the MSE value by dividing it with the average intensity of the ground truth observation, as recommended in \cite{zhong2022benchmarking, botev2021priors}. Additionally, we provide time history diagrams that plot the root mean square error (RMSE) against the normalized time, which maps the time interval $[T,2T]$ to the interval $[0,1]$. All evaluation metrics presented are averaged across all test trajectories and five runs, with mean and $75 \%$ inter-quartile range (IQR)  reported on all metrics.
Last, we also provide subsampled predictions and the pixelwise $L_2$ error of one trajectory for visual inspection but we emphasize  that one trajectory might not be representative for the overall performance (the trajectory shown is chosen at random). Please see Figure \ref{Fig: Training Testing Losses} for visual intuition on the training and testing procedure and how evaluation metrics are applied. 

\section{Results}
We report on a series of increasingly challenging experiments to test LaDID. First, we examined whether our model generalizes well on synthetic data for which the training and test 
data come from the same dynamical system.
This body of experiments test whether the
model can learn to map from a finite, empirical dataset to an effective latent dynamical model. Second, we examined few-shot generalization to 
data obtained from systems subject to nontrivial intervention (and in that sense strongly out-of-distribution).
In particular, we train our model on a set of trajectories under interventions, i.e. interventions upon the mass or length of the pendulum, changes to the Reynolds number, or variations to the camera view on the observed system, and apply the learned inductive bias to new and unseen interventional regimes in a few-shot learning setting.  This tests the hypothesis that the inductive bias of our learned latent dynamical models 
can be a useful proxy for dynamical systems exposed to a number of interventions. 

\subsection{Benchmark comparisons to state-of-the-art models for ODE and PDE problems }
We begin by investigating whether our approach can learn latent dynamical models in the 
conventional case in which the training and test 
data come from the same system.
We evaluate the performance of ODE-RNN, ODE2VAE, NODEP, MSVI and LaDID on data described in Sec. \ref{Sec: Datasets} and Sec. \ref{Appendix: Dataset details} of the Appendix with increasing order of difficulty, starting with the non-linear mechanical swing systems with underlying ODEs, before moving to non-linear cases based on PDEs (reaction-diffusion system, 2D wave equation, von K\'{a}rm\'{a}n vortex street at the transition from laminar to turbulent flows, and Naiver-Stokes equations). We only present results for a subset of performed experiments here 
but refer the interested reader to Appendix \ref{Appendix: Additional Results} for a detailed presentation of all results.

\begin{figure}[tbh]
\begin{subfigure}[h!]{0.44\textwidth}
         \centering
         \includegraphics[width=\textwidth]{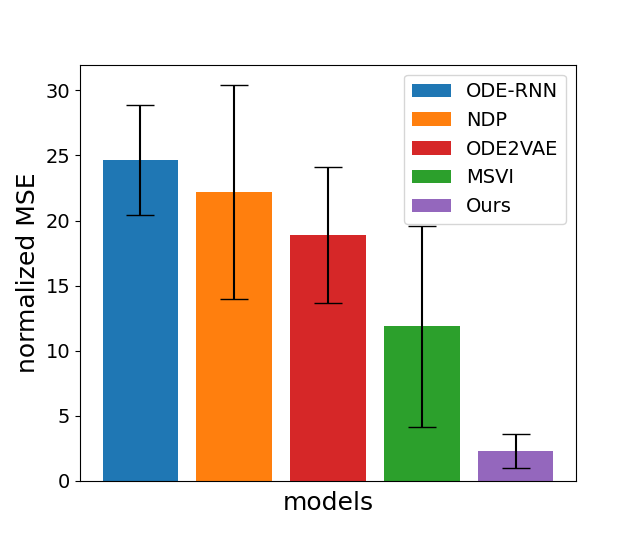}
         \caption{Normalized MSE }\label{Subfig: Single Pendulum MSE}
     \end{subfigure}
     \hfill
     \begin{subfigure}[h!]{0.55\textwidth}
         \centering
         \includegraphics[width=\textwidth]{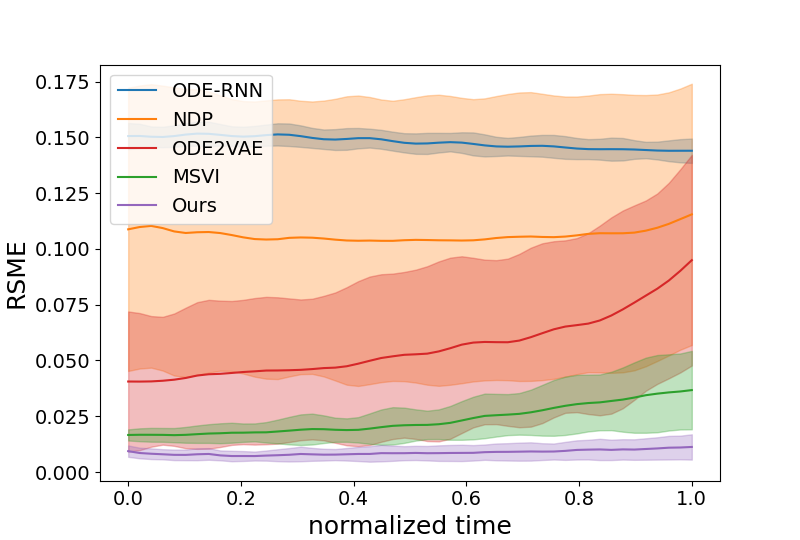}
         \caption{RMSE distribution over time}\label{Subfig: Single Pendulum RSME over time}
     \end{subfigure}
     \hfill
     \begin{subfigure}[h!]{0.99\textwidth}
         \centering
         \includegraphics[width=\textwidth]{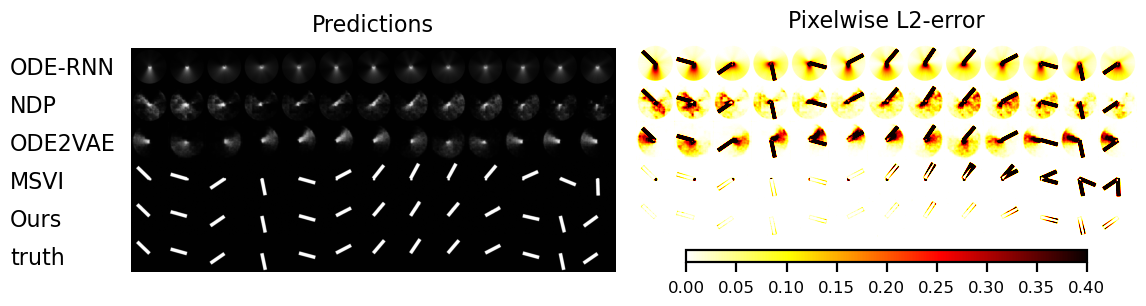}
         \caption{Example trajectory}\label{Subfig: Single Pendulum trajectory}
     \end{subfigure}
	 \caption{Test errors and exemplary test trajectories of different models for the single pendulum test case.}
  \label{Fig: single_pendulum_appendix}
\end{figure}

\textbf{Applications to ODE-based systems.} For visual inspection and intuition, Figure \ref{Subfig: Single Pendulum trajectory} provides predicted observations $\hat{x}_t^r$ of a few time points of one test trajectory of the single pendulum dataset for all tested algorithms, followed by the ground truth trajectory and the pixelwise $L_2$-error. 
In addition, Figure \ref{Subfig: Single Pendulum MSE} presents the normalized MSE over entire trajectories averaged across the entire test dataset and the evolution of the RMSE over time for the second half of the predicted observations averaged 
over all test trajectories (see Sec. \ref{Subsec: Metric}) is provided in Fig. \ref{Subfig: Single Pendulum RSME over time}. First, we see that across all ODE-based datasets LaDID 
achieves consistently the lowest normalized MSE. Second, the time history diagram (see Fig. \ref{Fig: single_pendulum_appendix} in the Appendix) shows  that LaDID predicts future time points with lower mean RMSE and lower standard deviation for long-horizon predictions relative to the other algorithms tested. 

\begin{figure}[tbh]
\begin{subfigure}[h!]{0.44\textwidth}
         \centering
         \includegraphics[width=\textwidth]{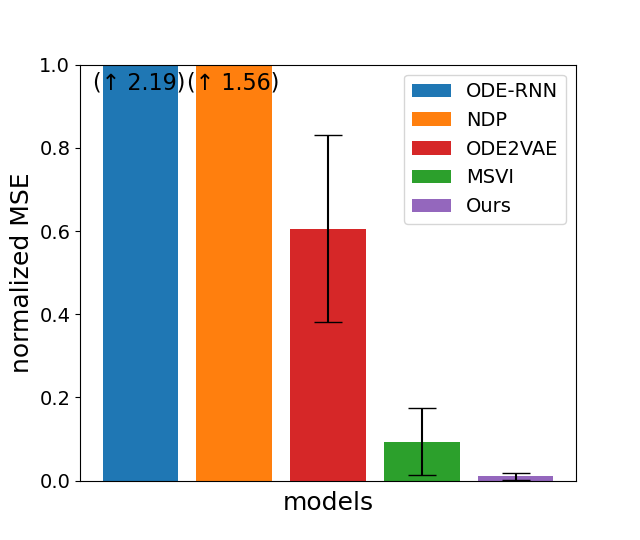}
         \caption{Normalized MSE }
     \end{subfigure}
     \hfill
     \begin{subfigure}[h!]{0.55\textwidth}
         \centering
         \includegraphics[width=\textwidth]{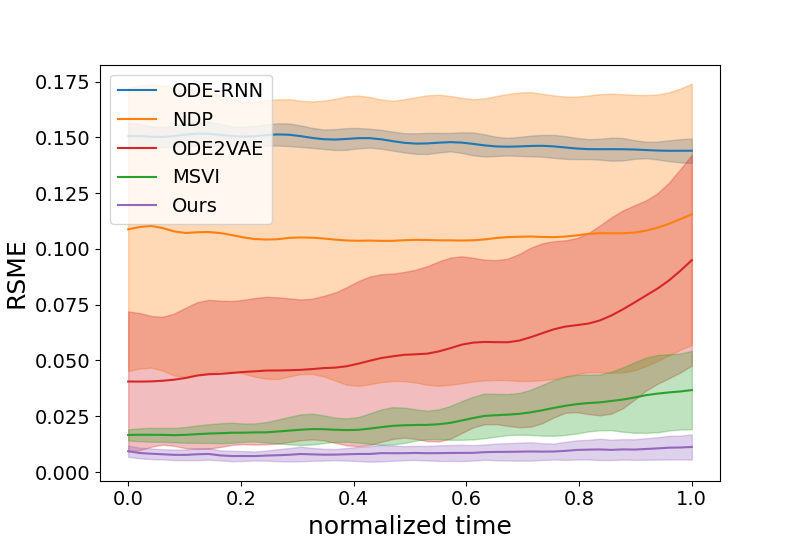}
         \caption{RMSE distribution over time}
     \end{subfigure}
     \hfill
     \begin{subfigure}[h!]{0.99\textwidth}
         \centering
         \includegraphics[width=\textwidth]{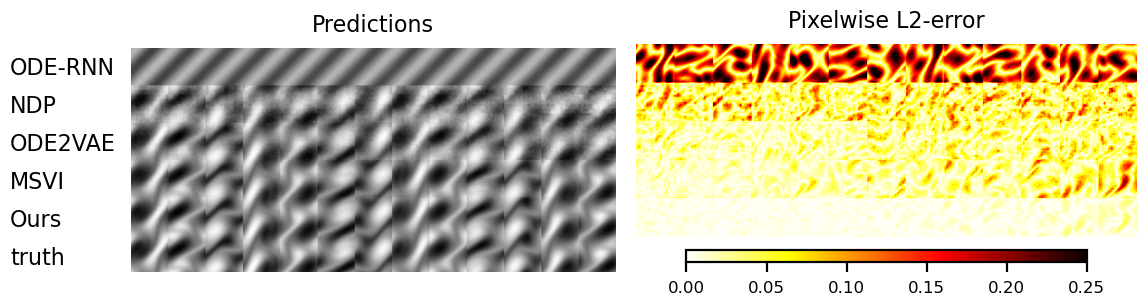}
         \caption{Example trajectory}
     \end{subfigure}
	 \caption{Test errors and exemplary test trajectories of different models for the Navier-Stokes equations test case.}
  \label{Fig: DINo_NS_appendix}
\end{figure}

This can also seen by visual inspection in Figure \ref{Subfig: Single Pendulum trajectory} as for other approaches the predicted states at later time points deviate from the ground truth trajectory substantially while LaDID's  predictions follow the ground truth. Considering only the baselines, one can observe that MSVI (a recent and sophisticated approach), achieves the best results and predicts accurately within a short-term horizon but fails on long-horizon predictions. The results for the challenging double pendulum test case can be found in the Appendix.

\textbf{Applications to PDE-based processes.} We additionally evaluated all baselines and our proposed method on PDE-based processes. Due to space restrictions, we focus our analysis on the flow evolution characterized by the Navier-Stokes equation in the two dimensional case, which is of  importance in many engineering tasks, e.g. the 
analysis of internal air flow in a combustion engine \cite{braun2019high, lagemann2022generalization},  drag reduction concepts in the transportation and energy sector \cite{gowree2018vortices, mateling2022analysis, mateling2023spanwise}, and many more. Results  in Figure \ref{Fig: DINo_NS_appendix} show that LaDID clearly outperforms all considered comparators in terms of normalized MSE and averaged RMSE. This is echoed in the other experiments whose results are presented in detail in Sec. \ref{Appendix: Additional Results} in the Appendix. 

Overall, the results shown (see also 
Sec. \ref{Appendix: Additional Results} for the complete collection of experimental results)
support the notion that LaDID can achieve  state-of-the-art performance for ODE and PDE based systems. 

\begin{figure}[h!]
\begin{subfigure}[h!]{0.49\textwidth}
         \centering
         \includegraphics[width=\textwidth]{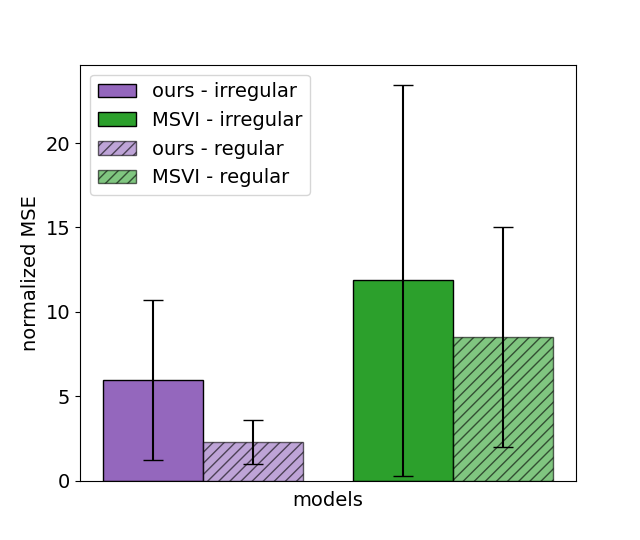}
         \caption{Single pendulum: Normalized MSE }
     \end{subfigure}
     \hfill
     \begin{subfigure}[h!]{0.5\textwidth}
         \centering
         \includegraphics[width=\textwidth]{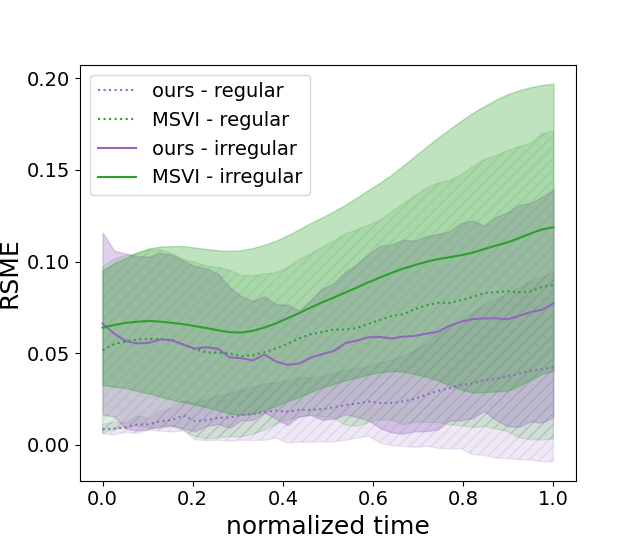}
         \caption{Sigle pendulum: RMSE distribution over time}
     \end{subfigure}
    \begin{subfigure}[h!]{0.49\textwidth}
         \centering
         \includegraphics[width=\textwidth]{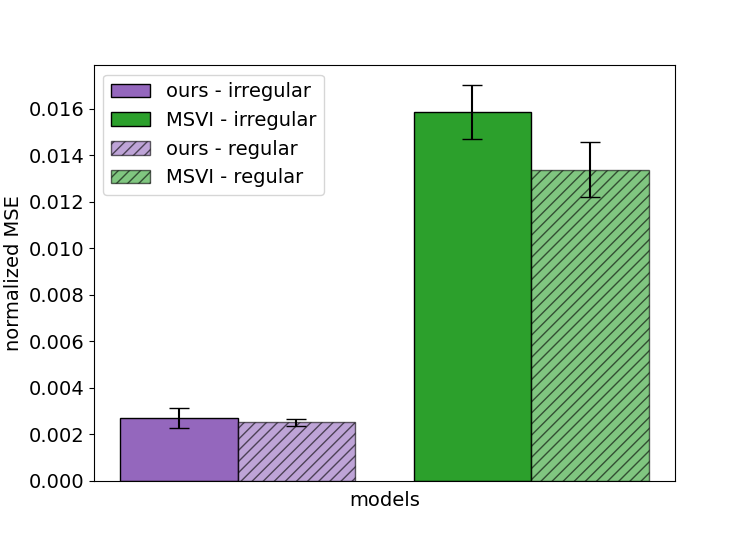}
         \caption{Lambda-omega reaction-diffusion system: Normalized MSE }
     \end{subfigure}
     \hfill
     \begin{subfigure}[h!]{0.5\textwidth}
         \centering
         \includegraphics[width=\textwidth]{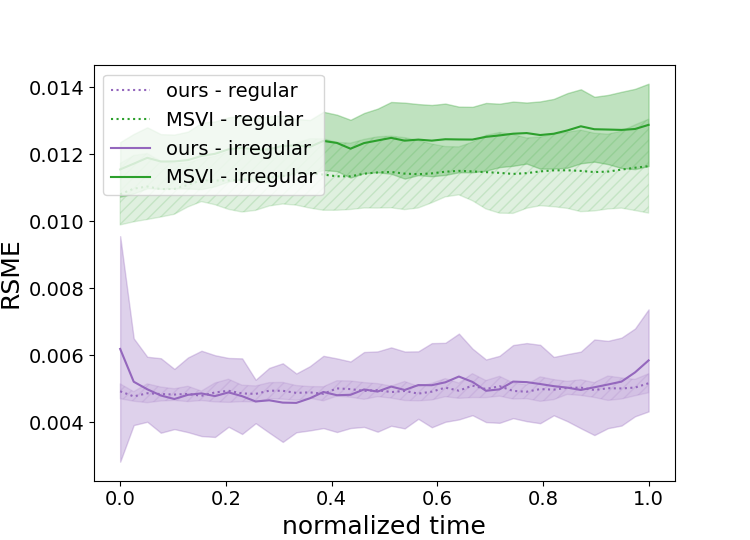}
         \caption{Lambda-omega reaction-diffusion systemRMSE distribution over time}
     \end{subfigure}
     
	 \caption{Test errors of different models for regular and irregular time grids: (a) Normalized MSE for single pendulum dataset, (b) RMSE over time for single pendulum dataset, (c) Normalized MSE for lambda-omega reaction-diffusion system, (d) RMSE over time for lambda-omega reaction-diffusion system.}
  \label{Fig: regular vs irregular}
\end{figure}

\textbf{Performance on regular and irregular time grids. } Here, we study the performance of LaDID on regular and irregular time grids and compare it to other 
neural-dynamical models (which are able to deal with irregular time series data). As shown in  Fig. \ref{Fig: regular vs irregular}, the proposed LaDID performs very similarly on both types of the time grids relative to both ODE-based benchmark examples and challenging PDE-based real-world systems, outperforming existing methods 
in these test cases with 
irregularly sampled data.

\begin{table}

\caption{Errors for ablated LaDID models trained on the single pendulum test case.}
\label{tab:ablations}
\resizebox{\textwidth}{!}{%
\renewcommand{\arraystretch}{0.8}
\begin{tabular}{ccc ccc ccc}
\hline
\multicolumn{3}{c}{\begin{tabular}[c]{@{}c@{}}loss \\ heuristics\end{tabular}}                                       & \multicolumn{3}{c}{\begin{tabular}[c]{@{}c@{}}attention\\ mechanism\end{tabular}}                                       & \multicolumn{3}{c}{\begin{tabular}[c]{@{}c@{}}representation\\ encoding\end{tabular}} \\ \hline
\multicolumn{1}{c}{ablation}                                             & \multicolumn{1}{c}{mean} & IQR           & \multicolumn{1}{c}{ablation}                                                & \multicolumn{1}{c}{mean} & IQR           & \multicolumn{1}{c}{ablation}  & \multicolumn{1}{c}{mean}  & \multicolumn{1}{c}{IQR} \\ \hline
reconstruction            & 2.66                      & 1.04          & no attention                                                                 & 7.79                      & 6.84          & w./o. encoding                 & 2.83                       & 2.02                     \\
\begin{tabular}[c]{@{}c@{}}reconstruction \& \\ representation\end{tabular} & 2.17                      & 0.99          & spatial attention                                                            & 2.81                      & 1.47          & \textbf{w. encoding}           & \textbf{2.02}              & \textbf{0.88}            \\
\begin{tabular}[c]{@{}c@{}}reconstruction \&\\ smoothness\end{tabular}      & 2.04                      & 0.92          & temporal attention                                                           & 2.41                      & 0.92          &                                &                            &                          \\
\textbf{full loss}                                                         & \textbf{2.02}             & \textbf{0.88} & \textbf{\begin{tabular}[c]{@{}c@{}}spatio-temporal\\ attention\end{tabular}} & \textbf{2.02}             & \textbf{0.88} &                                &                            &                         
\end{tabular}%
}
\end{table}
\textbf{Effects of relevant network modules.} As discussed above, our model leverages three key features: a reconstruction embedding, a spatio-temporal attention module and a specifically designed loss heuristic to learn temporal dynamics from empirical data. Here, we show that each of these network modules is indeed relevant for the  performance of LaDID (see Tab. \ref{tab:ablations}). First, we compare LaDID with counterparts trained on ablated loss heuristics, e.g. a pure reconstruction loss and loss combinations either using the described representation or smoothness loss. Overall, the proposed loss heuristic appears to stabilize training and yields the lowest MSE and IQR values.
Second, we compare LaDID to counterparts trained on ablated attention modules. Tab. \ref{tab:ablations} highlights that the applied spatio-temporal attention helps to extract key dynamical patterns. Finally, Tab. \ref{tab:ablations} further shows the usefulness of the proposed representation encoding. This representation encoding can be thought of a learning-enhanced initial value stabilizing the temporal evolution of latent trajectory dynamics. 

\subsection{Generalizing to novel systems via few-shot learning}
Here, we assess LaDID's ability to generalize to a
novel system obtained by nontrivial intervention
on the system coefficients themselves (e.g., mass, length, Reynolds number). 
Such changes can induce large changes to data distributions (and can also be viewed through a causal lens).
In this context, we aim to transfer the inductive bias learned from a set of training systems to a new set of systems with limited data availability in a few-shot learning setup.
In particular, we train a dynamical model on a set of interventions and fine-tune it to new intervention regimes with only a few samples, finally evaluating performance on an entirely unseen dataset. We compare the performance of our prior-based few-shot learning model with a model trained solely on the fine-tuning dataset (``scratch" trained model). In our first experiment, we use the single pendulum dataset and test the transferability hypothesis on fine-tuning datasets of varying sizes. The results show that the prior-based model outperforms the scratch-trained model at all fine-tuning dataset sizes, and achieves comparable performance to the model trained on the full dataset with a fine-tuning dataset size of $32\%$. At a fine-tuning dataset size of $8\%$, LaDID produces partially erroneous but still usable predictions, only slightly worse than the predictions of a recent NODE based model, MSVI,  trained on the full dataset.

\begin{figure}[h!]
\begin{subfigure}[h!]{0.49\textwidth}
         \centering
         \includegraphics[width=\textwidth, trim={0.5cm 0.75cm 1.6cm 2.0cm},clip]{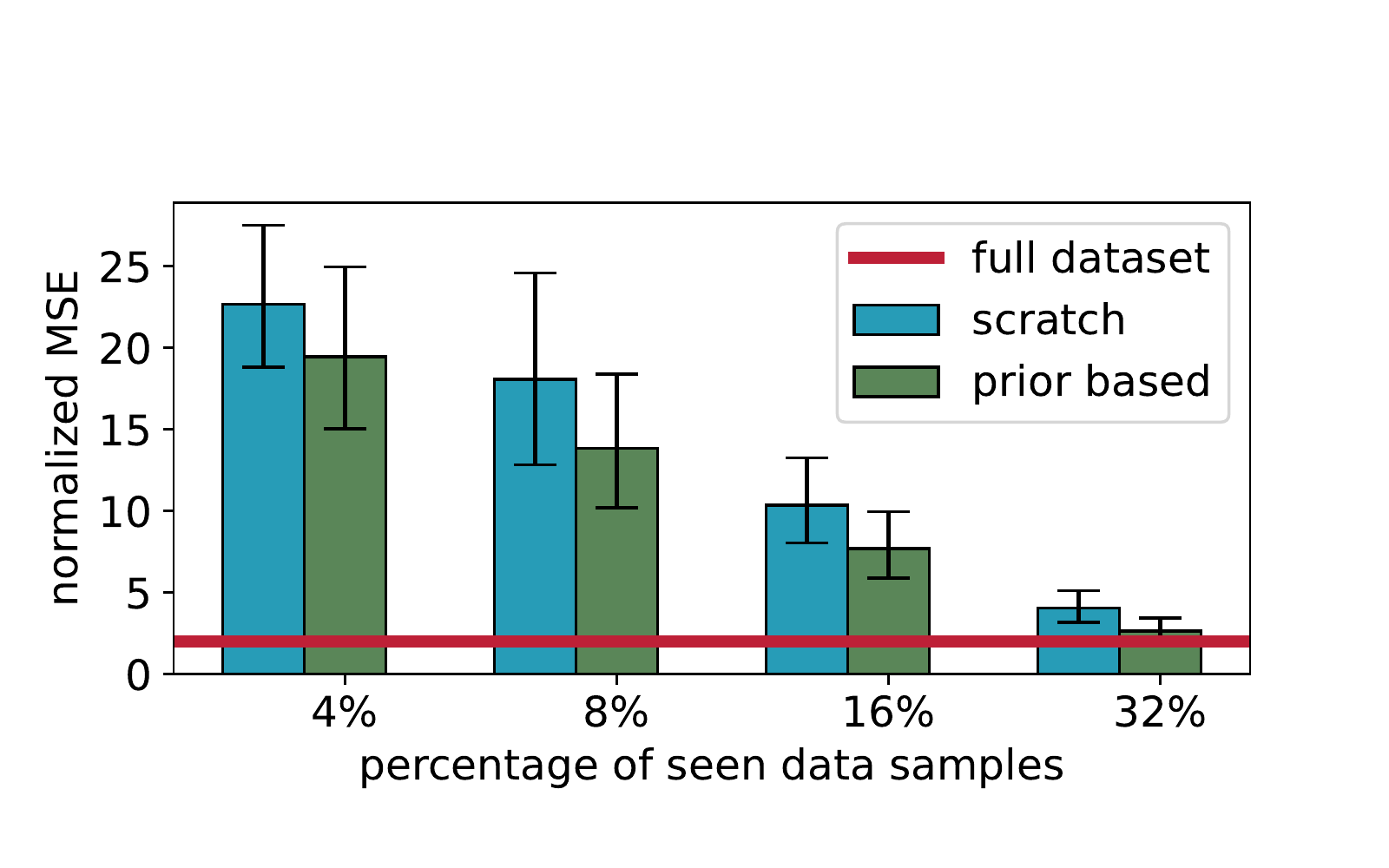}
         \caption{single pendulum}
     \end{subfigure}
     \hfill
     \begin{subfigure}[h!]{0.49\textwidth}
         \centering
         \includegraphics[width=\textwidth, trim={0.5cm 0.75cm 1.6cm 1.95cm},clip]{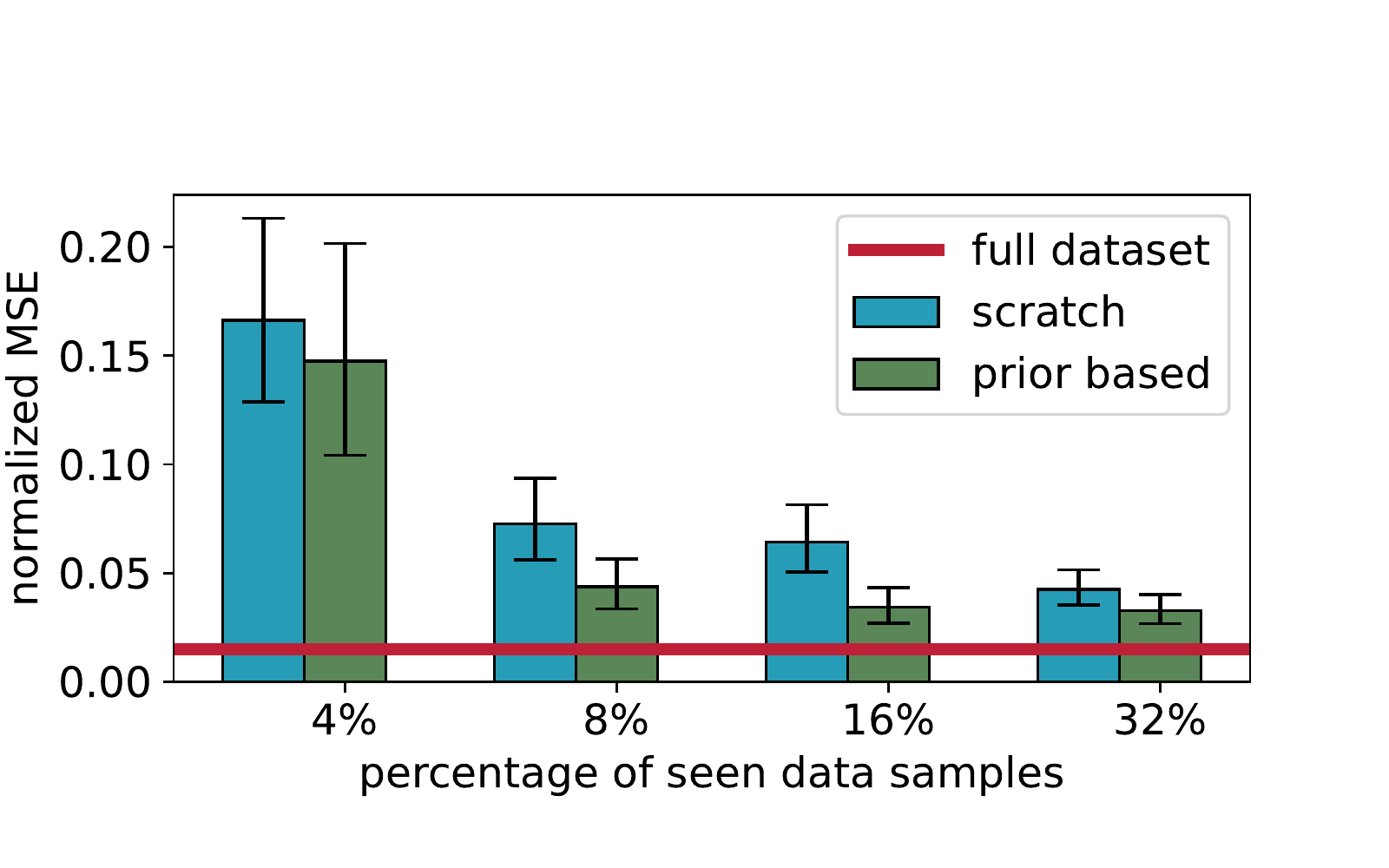}
         \caption{vortex street}
     \end{subfigure}
	 \caption{Test errors for a set of experiments involving transfer to nontrivial interventions (see text for details).}
  \label{Fig: fewShotLearning}
\end{figure}

In the second experiment, we investigate the effect of interventions on the observation process by testing the transferability of our model to new observation settings on the von K\'{a}rm\'{a}n vortex street dataset. We re-simulate different cylinder locations (shifted the cylinder to the left, right, up, and down) and evaluate performance with different fine-tuning dataset sizes. We find that the prior-based model outperforms the scratch-trained model and produces accurate and usable predictions under new conditions with a fine-tuning dataset size of as small as $8\%$. These findings support the notion that LaDID is capable of extracting generalizable dynamical models from training data.

\section{Discussion} \label{Sec: Discussion}

In this paper, we  presented a novel approach called LaDID aimed at end-to-end learning of latent dynamical models from empirical data. LaDID  
uses a novel transformer-based architecture 
that leverages certain scientifically-motivated invariances to allow
separation of a universal dynamics module and encoded realization-specific information.
We demonstated state-of-the-art performance on several new and challenging test cases and well-known benchmarks. Additionally, we showed that 
LaDID can generalize to systems under nontrivial intervention (when trained on the un-intervened system)
using few-shot learning, and in that sense provides a 
potentially useful way to model systems under novel interventions.

However, despite good performance relative to NODE-based methods, 
for complex models prediction over very long time horizons  remains challenging and 
further modifications and possibly  additional inductive bias will be needed for improved performance. At present, data sampled on an irregular \textit{spatial} grid cannot be considered. Future work using graph-based approaches will address this point.

Our work was motivated by the broad range of examples in contemporary science and engineering in which data are available from different experiments/pipelines that are plausibly underpinned by a unified model but where the model itself cannot be directly specified from first principles or is too complex to work with. 
A key motivation for our work is problems in the biomedical and health domains, where we expect the invariances underpinning LaDID to hold, but where explicit dynamical models are often not available at the outset. In ongoing work we are exploring the use of LaDID and related schemes in these challenging settings, e.g. for the modelling of complex longitudinal observations.
In such settings, an implicit yet universal model is useful as a way to capture system behaviour
and to generalize, in  a data efficient manner, to new realizations/instances where only limited data may be available.

\bibliographystyle{plain}
\bibliography{LaDID}

\begin{thebibliography}{10}

\bibitem{bai2019DeepEquilibriumModels}
Shaojie Bai, J~Zico Kolter, and Vladlen Koltun.
\newblock Deep equilibrium models.
\newblock {\em Advances in Neural Information Processing Systems}, 32, 2019.

\bibitem{benzi1992lattice}
Roberto Benzi, Sauro Succi, and Massimo Vergassola.
\newblock The lattice boltzmann equation: theory and applications.
\newblock {\em Physics Reports}, 222(3):145--197, 1992.

\bibitem{bhatnagar1954model}
Prabhu~Lal Bhatnagar, Eugene~P Gross, and Max Krook.
\newblock A model for collision processes in gases. i. small amplitude
  processes in charged and neutral one-component systems.
\newblock {\em Physical review}, 94(3):511, 1954.

\bibitem{botev2021priors}
Aleksandar Botev, Andrew Jaegle, Peter Wirnsberger, Daniel Hennes, and Irina
  Higgins.
\newblock Which priors matter? benchmarking models for learning latent
  dynamics.
\newblock {\em Advances in Neural Information Processing Systems}, 34, 2021.

\bibitem{braun2019high}
Marco Braun, Wolfgang Schr{\"o}der, and Michael Klaas.
\newblock High-speed tomographic {PIV} measurements in a {DISI} engine.
\newblock {\em Experiments in Fluids}, 60(9):146, 2019.

\bibitem{brunton2016DiscoveringGoverningEquationsFromData}
Steven~L. Brunton, Joshua~L. Proctor, and J.~Nathan Kutz.
\newblock Discovering governing equations from data by sparse identification of
  nonlinear dynamical systems.
\newblock {\em Proceedings of the National Academy of Sciences},
  113(15):3932--3937, 2016.

\bibitem{champion2019data}
Kathleen Champion, Bethany Lusch, J~Nathan Kutz, and Steven~L Brunton.
\newblock Data-driven discovery of coordinates and governing equations.
\newblock {\em Proceedings of the National Academy of Sciences},
  116(45):22445--22451, 2019.

\bibitem{chen2020learning}
Ricky T.~Q. Chen, Brandon Amos, and Maximilian Nickel.
\newblock Learning neural event functions for ordinary differential equations.
\newblock {\em International Conference on Learning Representations}, 2021.

\bibitem{chen2018NODE}
Ricky~TQ Chen, Yulia Rubanova, Jesse Bettencourt, and David~K Duvenaud.
\newblock Neural ordinary differential equations.
\newblock {\em Advances in Neural Information Processing Systems}, 31, 2018.

\bibitem{choi2022learning}
Matthew Choi, Daniel Flam-Shepherd, Thi~Ha Kyaw, and Al{\'a}n Aspuru-Guzik.
\newblock Learning quantum dynamics with latent neural ordinary differential
  equations.
\newblock {\em Physical Review A}, 105(4):042403, 2022.

\bibitem{cranmer2020LagrangianNeuralNetworks}
Miles Cranmer, Sam Greydanus, Stephan Hoyer, Peter Battaglia, David Spergel,
  and Shirley Ho.
\newblock Lagrangian neural networks.
\newblock {\em arXiv preprint arXiv:2003.04630}, 2020.

\bibitem{duong2021hamiltonian}
Thai Duong and Nikolay Atanasov.
\newblock {Hamiltonian-based Neural ODE Networks on the SE(3) Manifold For
  Dynamics Learning and Control}.
\newblock In {\em Proceedings of Robotics: Science and Systems}, 2021.

\bibitem{dupont2019AugmentedNODE}
Emilien Dupont, Arnaud Doucet, and Yee~Whye Teh.
\newblock Augmented neural odes.
\newblock {\em Advances in Neural Information Processing Systems}, 32, 2019.

\bibitem{Weinan2017MLviaDynamicalSystems}
Weinan E.
\newblock A proposal on machine learning via dynamical systems.
\newblock {\em Communications in Mathematics and Statistics}, 5(1):1--11, 2017.

\bibitem{fefferman2013ManifoldHypothesis}
Charles Fefferman, Sanjoy Mitter, and Hariharan Narayanan.
\newblock Testing the manifold hypothesis.
\newblock {\em Journal of the American Mathematical Society}, 29(4):983--1049,
  2016.

\bibitem{finlay2020train}
Chris Finlay, J{\"o}rn-Henrik Jacobsen, Levon Nurbekyan, and Adam Oberman.
\newblock How to train your neural ode: the world of jacobian and kinetic
  regularization.
\newblock In {\em International conference on machine learning}, pages
  3154--3164. PMLR, 2020.

\bibitem{finzi2020SimplifyingHNNLNNwithConstraints}
Marc Finzi, Ke~Alexander Wang, and Andrew~G Wilson.
\newblock Simplifying hamiltonian and lagrangian neural networks via explicit
  constraints.
\newblock {\em Advances in Neural Information Processing Systems}, 33, 2020.

\bibitem{gowree2018vortices}
Erwin~R Gowree, Chetan Jagadeesh, Edward Talboys, Christian Lagemann, and
  Christoph Br{\"u}cker.
\newblock Vortices enable the complex aerobatics of peregrine falcons.
\newblock {\em Communications biology}, 1(1):27, 2018.

\bibitem{greydanus2019HamiltonianNeuralNetworks}
Samuel Greydanus, Misko Dzamba, and Jason Yosinski.
\newblock Hamiltonian neural networks.
\newblock {\em Advances in Neural Information Processing Systems}, 32, 2019.

\bibitem{Haber2017StablArchsForNN}
Eldad Haber and Lars Ruthotto.
\newblock Stable architectures for deep neural networks.
\newblock {\em Inverse Problems}, 34(1):014004, 2017.

\bibitem{hanel2006molekulare}
Dieter H{\"a}nel.
\newblock {\em Molekulare Gasdynamik: Einf{\"u}hrung in die kinetische Theorie
  der Gase und Lattice-Boltzmann-Methoden}.
\newblock Springer-Verlag, 2006.

\bibitem{heinonen2018ODEwithGP}
Markus Heinonen, Cagatay Yildiz, Henrik Mannerstr{\"o}m, Jukka Intosalmi, and
  Harri L{\"a}hdesm{\"a}ki.
\newblock {Learning unknown ODE models with Gaussian processes}.
\newblock In {\em International Conference on Machine Learning}, pages
  1959--1968. PMLR, 2018.

\bibitem{iakovlev2022MSVI}
Valerii Iakovlev, Cagatay Yildiz, Markus Heinonen, and Harri
  L{\"a}hdesm{\"a}ki.
\newblock Latent neural {ODE}s with sparse bayesian multiple shooting.
\newblock In {\em The Eleventh International Conference on Learning
  Representations}, 2023.

\bibitem{jia2019NeuralJumpSDE}
Junteng Jia and Austin~R Benson.
\newblock Neural jump stochastic differential equations.
\newblock {\em Advances in Neural Information Processing Systems}, 32, 2019.

\bibitem{kaheman2020SINDyPi}
Kadierdan Kaheman, J.~Nathan Kutz, and Steven~L. Brunton.
\newblock {SINDy}-{PI}: a robust algorithm for parallel implicit sparse
  identification of nonlinear dynamics.
\newblock {\em Proceedings of the Royal Society A: Mathematical, Physical and
  Engineering Sciences}, 476(2242), 2020.

\bibitem{kidger2020CDEforIrregularTime}
Patrick Kidger, James Morrill, James Foster, and Terry Lyons.
\newblock Neural controlled differential equations for irregular time series.
\newblock {\em Advances in Neural Information Processing Systems}, 33, 2020.

\bibitem{kim2021inferring}
Timothy~D Kim, Thomas~Z Luo, Jonathan~W Pillow, and Carlos~D Brody.
\newblock Inferring latent dynamics underlying neural population activity via
  neural differential equations.
\newblock In {\em International Conference on Machine Learning}, pages
  5551--5561. PMLR, 2021.

\bibitem{kingma2013autoencoding}
Diederik~P Kingma and Max Welling.
\newblock Auto-encoding variational bayes.
\newblock {\em arXiv preprint arXiv:1312.6114}, 2013.

\bibitem{lagemann2022generalization}
Christian Lagemann, Kai Lagemann, Sach Mukherjee, and Wolfgang Schroeder.
\newblock Generalization of deep recurrent optical flow estimation for
  particle-image velocimetry data.
\newblock {\em Measurement Science and Technology}, 2022.

\bibitem{li2020FourierNeuralOperator}
Zongyi Li, Nikola Kovachki, Kamyar Azizzadenesheli, Burigede Liu, Kaushik
  Bhattacharya, Andrew Stuart, and Anima Anandkumar.
\newblock Fourier neural operator for parametric partial differential
  equations.
\newblock In {\em The Eighth International Conference on Learning
  Representations}, 2020.

\bibitem{loshchilov2019decoupled}
Ilya Loshchilov and Frank Hutter.
\newblock Decoupled weight decay regularization.
\newblock {\em arXiv preprint arXiv:1711.05101}, 2017.

\bibitem{lu2020BeyondFiniteLayerNN}
Yiping Lu, Aoxiao Zhong, Quanzheng Li, and Bin Dong.
\newblock Beyond finite layer neural networks: Bridging deep architectures and
  numerical differential equations.
\newblock In {\em International Conference on Machine Learning}, pages
  3276--3285. PMLR, 2018.

\bibitem{lutter2019DeepLagrangianNets}
Michael Lutter, Christian Ritter, and Jan Peters.
\newblock Deep lagrangian networks: Using physics as model prior for deep
  learning.
\newblock In {\em International Conference on Learning Representations}, 2019.

\bibitem{massaroli2021differentiableMultipleShootingLayers}
Stefano Massaroli, Michael Poli, Sho Sonoda, Taiji Suzuki, Jinkyoo Park,
  Atsushi Yamashita, and Hajime Asama.
\newblock Differentiable multiple shooting layers.
\newblock {\em Advances in Neural Information Processing Systems},
  34:16532--16544, 2021.

\bibitem{mateling2023spanwise}
Esther M{\"a}teling, Marian Albers, and Wolfgang Schr{\"o}der.
\newblock How spanwise travelling transversal surface waves change the
  near-wall flow.
\newblock {\em Journal of Fluid Mechanics}, 957:A30, 2023.

\bibitem{mateling2022analysis}
Esther M{\"a}teling and Wolfgang Schr{\"o}der.
\newblock Analysis of spatiotemporal inner-outer large-scale interactions in
  turbulent channel flow by multivariate empirical mode decomposition.
\newblock {\em Physical Review Fluids}, 7(3):034603, 2022.

\bibitem{morill2020NeuralRoughODEForLongTime}
James Morrill, Cristopher Salvi, Patrick Kidger, and James Foster.
\newblock Neural rough differential equations for long time series.
\newblock In {\em International Conference on Machine Learning}, pages
  7829--7838. PMLR, 2021.

\bibitem{paszke2019pytorch}
Adam Paszke, Sam Gross, Francisco Massa, Adam Lerer, James Bradbury, Gregory
  Chanan, Trevor Killeen, Zeming Lin, Natalia Gimelshein, Luca Antiga, Alban
  Desmaison, Andreas Köpf, Edward Yang, Zach DeVito, Martin Raison, Alykhan
  Tejani, Sasank Chilamkurthy, Benoit Steiner, Lu~Fang, Junjie Bai, and Soumith
  Chintala.
\newblock Pytorch: An imperative style, high-performance deep learning library,
  2019.

\bibitem{qian1992lattice}
Yue-Hong Qian, Dominique d'Humi{\`e}res, and Pierre Lallemand.
\newblock {Lattice BGK models for Navier-Stokes equation}.
\newblock {\em Europhysics letters}, 17(6):479, 1992.

\bibitem{rezende2014stochastic}
Danilo~Jimenez Rezende, Shakir Mohamed, and Daan Wierstra.
\newblock Stochastic backpropagation and approximate inference in deep
  generative models.
\newblock In {\em International conference on machine learning}, pages
  1278--1286. PMLR, 2014.

\bibitem{richterpowell2022neuralConservationLaws}
Jack Richter-Powell, Yaron Lipman, and Ricky T.~Q. Chen.
\newblock {Neural Conservation Laws: A Divergence-Free Perspective}.
\newblock In {\em Advances in Neural Information Processing Systems},
  volume~35, 2022.

\bibitem{rubanova2019ODERNN}
Yulia Rubanova, Ricky T.~Q. Chen, and David~K Duvenaud.
\newblock Latent ordinary differential equations for irregularly-sampled time
  series.
\newblock In {\em Advances in Neural Information Processing Systems},
  volume~32, 2019.

\bibitem{ruthotto2018NNbyPDE}
Lars Ruthotto and Eldad Haber.
\newblock Deep neural networks motivated by partial differential equations.
\newblock {\em Journal of Mathematical Imaging and Vision}, 62:352--364, 2020.

\bibitem{yıldız2019ode2vae}
Cagatay Yildiz, Markus Heinonen, and Harri Lahdesmaki.
\newblock {ODE\textsuperscript{2}VAE: Deep generative second order ODEs with
  Bayesian neural networks}.
\newblock In {\em Advances in Neural Information Processing Systems},
  volume~32, 2019.

\bibitem{Yin2023}
Yuan Yin, Matthieu Kirchmeyer, Jean-Yves Franceschi, Alain Rakotomamonjy, and
  Patrick Gallinari.
\newblock {Continuous PDE Dynamics Forecasting with Implicit Neural
  Representations}.
\newblock In {\em International Conference on Learning Representations}, 2023.

\bibitem{zhi2022learning}
Weiming Zhi, Tin Lai, Lionel Ott, Edwin~V Bonilla, and Fabio Ramos.
\newblock Learning efficient and robust ordinary differential equations via
  invertible neural networks.
\newblock In {\em International Conference on Machine Learning}, pages
  27060--27074. PMLR, 2022.

\bibitem{zhong2020SympleticODENet-HNNwithControl}
Yaofeng~Desmond Zhong, Biswadip Dey, and Amit Chakraborty.
\newblock {Symplectic ODE-Net: Learning Hamiltonian Dynamics with Control}.
\newblock In {\em International Conference on Learning Representations}, 2020.

\bibitem{zhong2022benchmarking}
Yaofeng~Desmond Zhong, Biswadip Dey, and Amit Chakraborty.
\newblock Benchmarking energy-conserving neural networks for learning dynamics
  from data.
\newblock In {\em Proceedings of the 3rd Conference on Learning for Dynamics
  and Control}, pages 1218--1229. PMLR, 2021.

\end{thebibliography}

\FloatBarrier
\newpage
\appendix

\section{Training details and hyperparameters} \label{Appendix: Training details}
Our implementation uses the PyTorch framework \cite{paszke2019pytorch}. All modules are initialized from scratch using random weights. During training, an AdamW-Optimizer \cite{loshchilov2019decoupled} is applied starting at an initial learning rate $\varepsilon_0=0.0003$. An exponential learning rate scheduler is applied showing the best results in the current study. Every network is trained for 30 000 epochs. At initialization, we start training at a subpatch length of 1 which is doubled every 3000 epochs. After the CNN encoder, 8 attention layers are stacked each using 4 attention heads. A relative Sin/Cos embedding is used as position encoding followed by a linear layer. The input resolution of the observational image data is $128 \times 128$ px. All computations are run on a single GPU node equipped with one NVidia A100 (40 GB) and a global batch size of 16 is used. A full training run on the single pendulum, the double pendulum, the wave equation and the Navier-Stokes equation dataset requires approx. 14 h. A full training run on the reaction-diffusion system and the von K\'{a}rm\'{a}n vortex street requires approx. 8 h. 

\begin{table}[h!]
    \centering
    \caption{Training hyperparameters}
  \label{Tab: hyperparameters}
    \begin{tabular}{l l}

        Hyperparameter & Value \\ \hline
        LR schedule & Exp. decay \\ 
        Initial LR & 3e-4 \\ 
        Weight Decay & 0.01 \\ 
        Global batch size & 16 \\ 
        Parallel GPUs & 1 \\ 
        Input resolution & $128 \times 128$ px \\ 
        Number of input timesteps & 10 \\ 
        Initial subpatch length & 1 \\
        Number of epochs per subpatch length & 3000 \\
        Latent dimension & 32 \\
        attention mechanism & spatio-temporal \\ 
        Number of attention blocks & 8 \\ 
        Number of attention heads & 4 \\ 
        Position Encoding & relative Sin/Cos encoding \\
    \end{tabular}
\end{table}

\section{Dataset details} \label{Appendix: Dataset details}
Here, we provide details about the datasets used in this work, their underlying mathematical formulations and their implementation details. Partially, some of the datasets we selected are used in literature to demonstrate the effectiveness of neural based temporal modeling approaches, e.g., a swinging pendulum or a reaction-diffusion system. However, we also consider more unknown test cases which represent complex real-world applications such as the chaotic double-pendulum or fluid flow applications driven by a complex set of PDEs, i.e., the Navier-Stokes equations. 

\subsection{Swinging pendulum}
For the first dataset we consider synthetic videos of a nonlinear pendulum simulated in two spatial dimensions. Typically, a nonlinear swinging pendulum is governed by the following second order differential equation:
\begin{equation}
    \frac{d^2 z}{dt^2} = - \sin z
\end{equation}
with $z$ denoting the angle of the pendulum. Overall, we simulated 500 trajectories with different initial conditions. For each trajectory, the initial angle $z$ and its angular velocity $\frac{d z}{dt}$ is sampled uniformly from $z \sim \mathcal{U}(0,2 \pi)$ and $\frac{d z}{dt} \sim \mathcal{U}(-\pi / 2, \pi/2)$. All trajectories are simulated for $t=3$ seconds. The training, validation and test dataset is split into 400, 50 and 50 trajectories, respectively. The swinging pendulum is rendered in black/white image space over 128 pixels for each spatial dimension. Hence, each observation is a high-dimensional image representation (16384 dimensions - flattened $128 \times 128 \, \textrm{px}^2$ image) of an instantaneous state of the second-order ODE. 

\subsection{Swinging double pendulum}
To increase the complexity of the second dataset, we selected the kinematics of a nonlinear double pendulum motion. The pendulums are treated as two point masses with the upper pendulum being denoted by the subscript "1" and the lower one by subscript "2". The kinematics of this nearly chaotic system is governed by the following set of ordinary differential equations: 
{
\small
\begin{align}
 \frac{d^2 z_1}{dt^2} = \frac{-g(2 m_1 + m_2)\sin z_1 - m_2 g \sin ( z_1 -2 z_2) - 2 \sin (z_1 -z_2) m_2 (\frac{dz_1}{dt}^2 L_2 + \frac{dz_1}{dt}^2 L_1 \cos (z_1 - z_2))}{L_1(2m_1 + m_2 - m_2 \cos (2 z_1 - 2 z_2))}\\
 \frac{d^2 z_2}{dt^2} = \frac{2 \sin ( z_1 - z_2)(\frac{dz_1}{dt}^2 L_1 (m_1 + m_2)+ g (m_1 + m_2) \cos z_1 + \frac{dz_2}{dt}^2 L_2 m_2 \cos (z_1 - z_2))}{L_2(2m_1 + m_2 - m_2 \cos (2 z_1 - 2 z_2))}
\end{align}
} %

with $m_i$ denoting the mass and the length of each pendulum respectively, and $g$ is the gravitational constant. Again, we simulated 500 trajectories split in sets of 400, 50 and 50 samples for training, validation and testing. The initial condition for $(z_1, z_2)$ and $(\frac{dz_1}{dt}, \frac{dz_2}{dt})$ are uniformly sampled in the range $\mathcal{U}(0, 2 \pi)$ and $\mathcal{U}(-\pi / 2, \pi/2)$. The double pendulum is rendered in a RGB color space over 128 pixels for each spatial dimension with the first pendulum colored in red and the second one in green. Hence, each observation is a high-dimensional image representation ($16384 \times 3$ dimensions - flatted $128 \times 128 \, \textrm{px}^2$ RGB image) of an instantaneous double pendulum state. 

\subsection{Reaction-diffusion equation}
Many real-world applications of interest originate from dynamics governed by partial differential equations with more complex interactions between spatial and temporal dynamics. One such set of PDEs we selected as test case is based on a lambda-omega reaction-diffusion system which is described by the following equations: 
\begin{align}
    \frac{du}{dt} = (1 -(u^2 + v^2))u + \beta (u^2 + v^2) v + d_1 (\frac{d^2 u}{dx^2} + \frac{d^2 u}{dy^2})\\
    \frac{dv}{dt} = -\beta (u^2 + v^2)u + (1- (u^2 + v^2)) v + d_2 (\frac{d^2 v}{dx^2} + \frac{d^2 v}{dy^2})
\end{align}
with $(d_1, d_2) = 0.1$ denoting diffusion constants and $\beta=1$. This set of equations generates a spiral wave formation which can be approximated by two oscillating spiral modes. The system is simulated from a single initial condition from $t=0$ to $t=10$ in $\Delta t=0.05$ time intervals for a total number of 10 000 samples. The initial conditions is defined as 
\begin{align}
    u(x,y,0) = \mathrm{tanh} \left ( \sqrt{x^2 + y^2} \cos \left ( (x + i y) - \sqrt{x^2 - y^2}\right ) \right )\\
    v(x,y,0) = \mathrm{tanh} \left ( \sqrt{x^2 + y^2} \sin \left ( (x + i y) - \sqrt{x^2 - y^2}\right )\right ).
\end{align}
The simulation is performed over a spatial domain of $(x \in [-10, 10]$ and $y \in [-10, 10]$ on grid with 128 points in each spatial dimension. We split this simulation into trajectories of 50 consecutive samples resulting in 200 in dependant realisations. We use 160 randomly sampled trajectories for training, 20 trajectories for validation and the remaining 20 trajectories for testing. Source code of the simulation can be found in \cite{champion2019data}.

\subsection{Two-dimensional wave equation}
A classical example of a hyperbolic PDE is the two-dimensional wave equation describing the temporal and spatial propagation of waves such as sound or water waves. Wave equations are important for a variety of fields including acoustics, electromagnetics and fluid dynamics. In two dimensions, the wave equation can be described as follows:
 \begin{equation}
     \frac{\partial ^2 u}{\partial t^2} = c^2 \triangledown ^2 u,
 \end{equation}
with $\triangledown^2$ denoting the Laplacian operator in $\mathcal{R}^2$ and $c$ is a constant speed of the wave propagation. The initial displacement $u_0$ is a Gaussian function 
\begin{equation}
    u_0 = a \exp \left(- \frac{(x-b)^2}{ 2r^2} \right ),
\end{equation}
where the amplitude of the peak displacement $a$, the location of the peak displacement $b$ and the standard deviation $r$ are uniformly sampled from $a \sim \mathcal{U}(2, 4)$, $b \sim \mathcal{U}(-1, 1)$, and $r \sim \mathcal{U}(0.25, 0.30)$, respectively. Similar to \cite{Yin2023}, the inital velocity gradient $\frac{\partial u}{\partial t}$ is set to zero. The wave simulations are performed over a spatial domain of $(x \in [-1, 1]$ and $y \in [-1, 1]$ on a grid with 128 points in each spatial dimension. Overall, 500 independent trajectories (individual initial conditions) are computed which are split in 400 randomly sampled trajectories for training, 50 trajectories for validation and the remaining 50 trajectories for testing.

\subsection{Navier-Stokes equations}
To ultimately test the performance of our model on complex real-world data, we simulated fluid flows governed by a complex set of partial differential equations called Navier-Stokes equations. Overall, two flow cases of different nature are considered, e.g., the temporal evolution of generic initial vorticity fields and the flow around an obstacle characterized by the formations of dominant vortex patterns also known as the von K\'{a}rm\'{a}n vortex street. \\
Due to the characteristics of the selected flow fields, we consider the incompressible two-dimensional Navier-Stokes equations given by 
\begin{equation}
    \frac{\partial u}{\partial t} + (u \cdot \triangledown)u - \nu \triangledown^2 u = - \frac{1}{\rho} \triangledown p .
\end{equation}
Here, $u$ denotes the velocity in two dimensions, $t$ and $p$ are the time and pressure, and $\nu$ is the kinematic viscosity. For the generic test case, we solve this set of PDEs in its vorticity form and chose initial conditions as described in \cite{li2020FourierNeuralOperator}. Simulations are performed over a spatial domain of $(x \in [-1, 1]$ and $y \in [-1, 1]$ on a grid with 128 points in each spatial dimension. Overall, 500 independent trajectories (individual initial vorticity fields) are computed which are split in 400 randomly sampled trajectories for training, 50 trajectories for validation and the remaining 50 trajectories for testing. \\

\subsection{Flow around a blunt body}
The second fluid flow case mimics an engineering inspired applications and captures the flow around a blunt cylinder body, also known as von K\'{a}rm\'{a}n vortex street. von K\'{a}rm\'{a}n vortices manifest in a repeating pattern of swirling vortices caused by the unsteady flow separation around blunt bodies and occur when the inertial forces in a flow are significantly greater than the viscous forces. A large dynamic viscosity  of a fluid suppresses vortices, whereas a higher density, velocity, and larger size  of the flowed object provide for more dynamics and a less ordered flow pattern. If the factors that increase the inertial forces are put in relation to the viscosity, a dimensionless measure - the Reynolds number -  is obtained that can be used to characterize a flow regime. If the Reynolds number is larger than $Re > 80$, the two vortices in the wake of the body become unstable until they finally detach periodically. The detached vortices remain stable for a while until they slowly dissociate again in the flow due to friction, and finally disappear. The incompressible vortex street is simulated using an open-source Lattice-Boltzmann solver due to computational efficiency. The governing equation is the Boltzmann equation with the simplified right-hand side (RHS) Bhatnagar-Gross-Krook (BGK) collision term \cite{bhatnagar1954model}:
\begin{equation}
    \frac{\partial f}{\partial t} + \zeta_k \frac{\partial f}{\partial x_k} = - \frac{1}{\tau} (f- f^{eq})
    \label{eq:lbm}
\end{equation}
These particle probability density functions (PPDFs) $f = f(\vec{x}, \vec{\zeta}, t)$ describe the probability to find a fluid particle around a location $\vec{x}$ with a particle velocity $\vec{\zeta}$ at time $t$ \cite{benzi1992lattice}. The left-hand side (LHS) describes the evolution of fluid particles in space and time, while the RHS describes the collision of particles. The collision process is governed by the relaxation parameter $1/\tau$ with the relaxation time $\tau$ to reach the Maxwellian equilibrium state $f^{eq}$. The discretized form of equation \ref{eq:lbm} yield the lattice-BGK equation 
\begin{equation}
    f_k(\vec{x} + \zeta_k \Delta t, t + \Delta t) = f_k (\vec{x}, t) - \frac{1}{\tau} ( f_k (\vec{x}, t) - f_k^{eq} (\vec{x}, t)).
\end{equation}
The standard $D_2Q_9$ discretization scheme with nine PPDFs \cite{qian1992lattice} is applied. The equilibrium PPDF is given by 
\begin{equation}
f_k ^{eq} = w_k \rho \left ( 1 + \frac{\zeta_k \vec{u}}{c_s^2} + \frac{(\zeta_k \vec{u})^2}{2c_s^4} - \frac{ \vec{u}^2}{2c_s^2}\right )
\end{equation}
where the quantities $w_k$ are weighting factors for the $D_2Q_9$ scheme given by $4/9$ for $k \in {0}$, $1/9$ for $k \in {1, \dots, 4}$, and $1/36$ for $k \in {5, \dots 9}$, and $\vec{u}$ is the fluid velocity. $c_s$ denotes the speed of sound. The makroscopic variables can be obtained from the moments of the PPDFs. Within the continuum limit, i.e., for small Knudsen numbers, the Navier-Stokes equations can directly be derived from the Boltzmann equation and the BGK model \cite{hanel2006molekulare}. We simulated three different Reynolds numbers $Re=100, 250, 500$ for 425 000 iterations with a mesh size of 128 point in vertical and 256 points in horizontal direction. We skipped the first 25 000 iterations to ensure a developed flow field and extracted velocity snapshot every 100 iterations. The simulation is performed over a spatial domain of $(x \in [-20, 20]$ and $y \in [-10, 10]$. We split this simulation into trajectories of 50 consecutive samples resulting in 200 in dependant realisations. We use 160 randomly sampled trajectories for training, 20 trajectories for validation and the remaining 20 trajectories for testing.\\

\newpage

\section{Additional results} \label{Appendix: Additional Results}
\counterwithin{figure}{section}
\setcounter{figure}{0}  

\subsection{Swinging double pendulum}
\begin{figure}[h!]
\begin{subfigure}[h!]{0.49\textwidth}
         \centering
         \includegraphics[width=\textwidth]{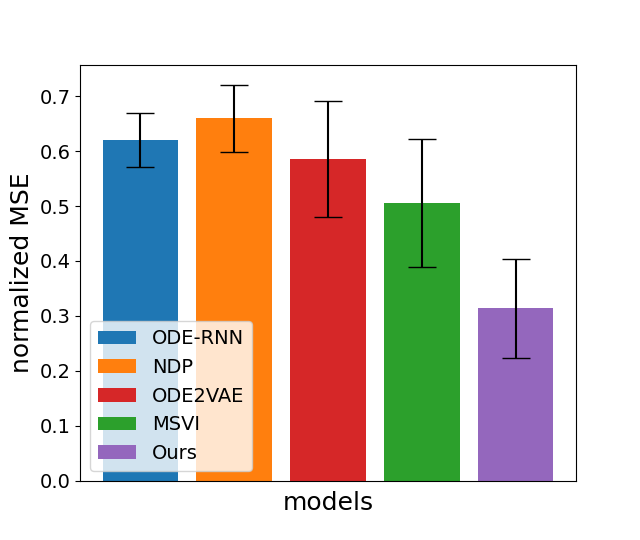}
         \caption{Normalized MSE }
     \end{subfigure}
     \hfill
     \begin{subfigure}[h!]{0.5\textwidth}
         \centering
         \includegraphics[width=\textwidth]{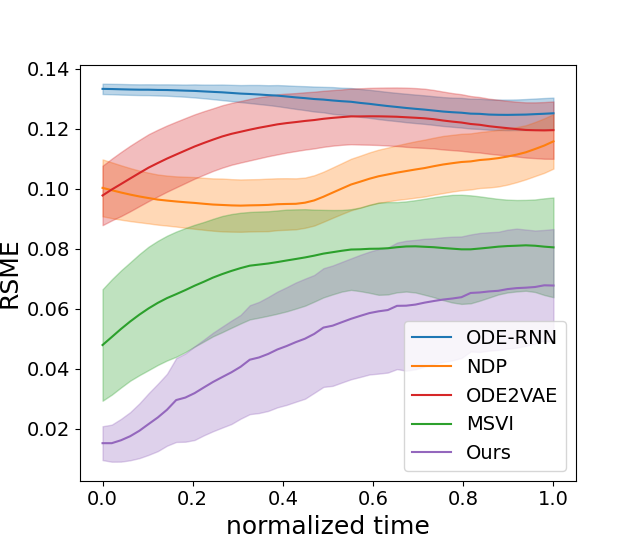}
         \caption{RMSE distribution over time}
     \end{subfigure}
     \hfill
     \begin{subfigure}[h!]{0.99\textwidth}
         \centering
         \includegraphics[width=\textwidth]{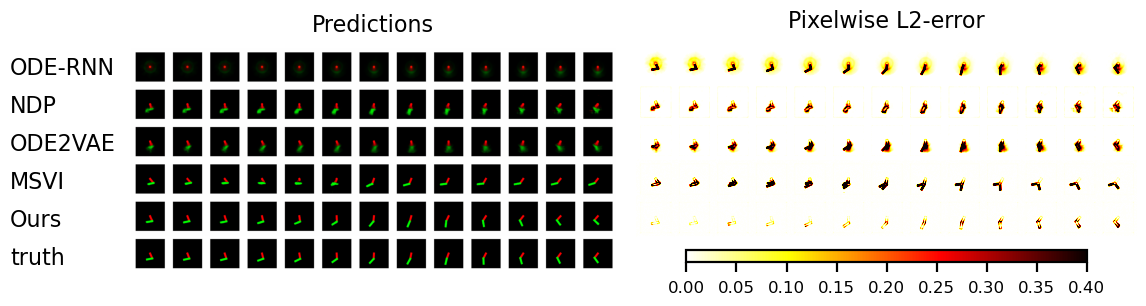}
         \caption{Example trajectory}
     \end{subfigure}
	 \caption{Test errors and exemplary test trajectories of different models for the double pendulum test case.}
  \label{Fig: double_pendulum_appendix}
\end{figure}
\FloatBarrier
\newpage
\subsection{Lambda-omega reaction-diffusion system}
\begin{figure}[h!]
\begin{subfigure}[h!]{0.51\textwidth}
         \centering
         \includegraphics[width=\textwidth]{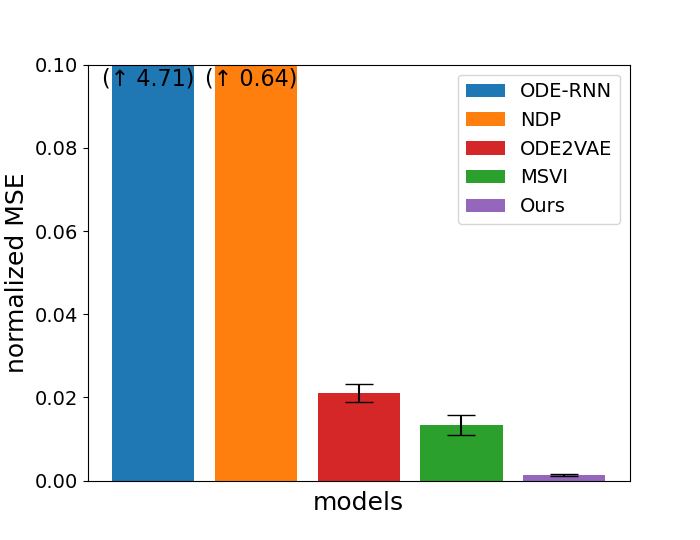}
         \caption{Normalized MSE }
     \end{subfigure}
     \hfill
     \begin{subfigure}[h!]{0.47\textwidth}
         \centering
         \includegraphics[width=\textwidth]{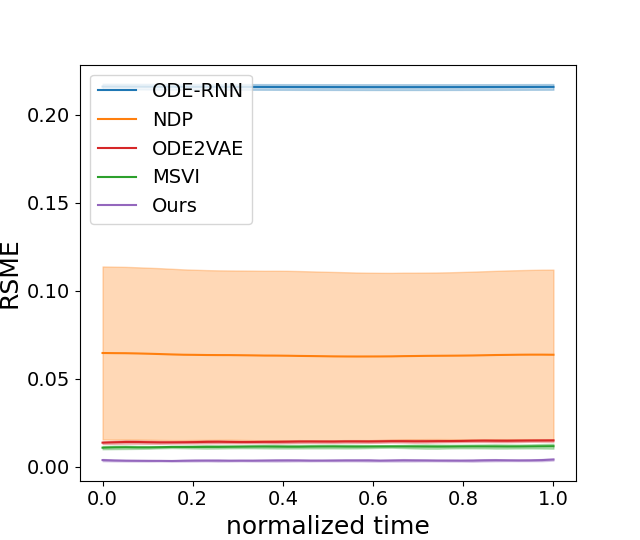}
         \caption{RMSE distribution over time}
     \end{subfigure}
     \hfill
     \begin{subfigure}[h!]{0.99\textwidth}
         \centering
         \includegraphics[width=\textwidth]{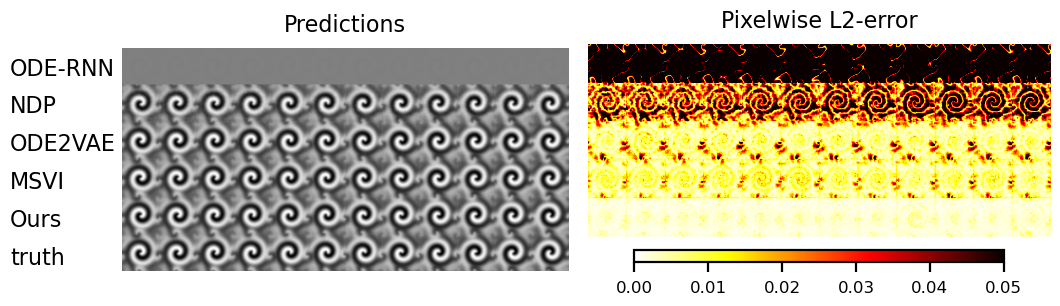}
         \caption{Example trajectory}
     \end{subfigure}
	 \caption{Test errors and exemplary test trajectories of different models for the lambda-omega reaction-diffusion system.}
  \label{Fig: reaction_diffusion_appendix}
\end{figure}

\FloatBarrier
\newpage
\subsection{Two-dimensional wave equation}
\begin{figure}[h!]
\begin{subfigure}[h!]{0.44\textwidth}
         \centering
         \includegraphics[width=\textwidth]{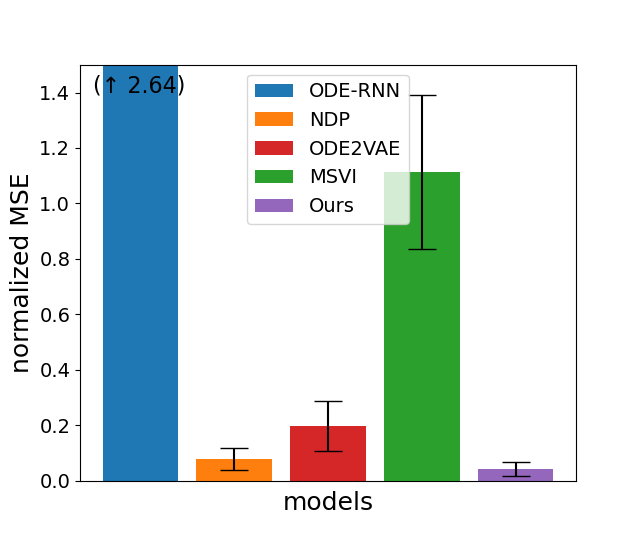}
         \caption{Normalized MSE }
     \end{subfigure}
     \hfill
     \begin{subfigure}[h!]{0.55\textwidth}
         \centering
         \includegraphics[width=\textwidth]{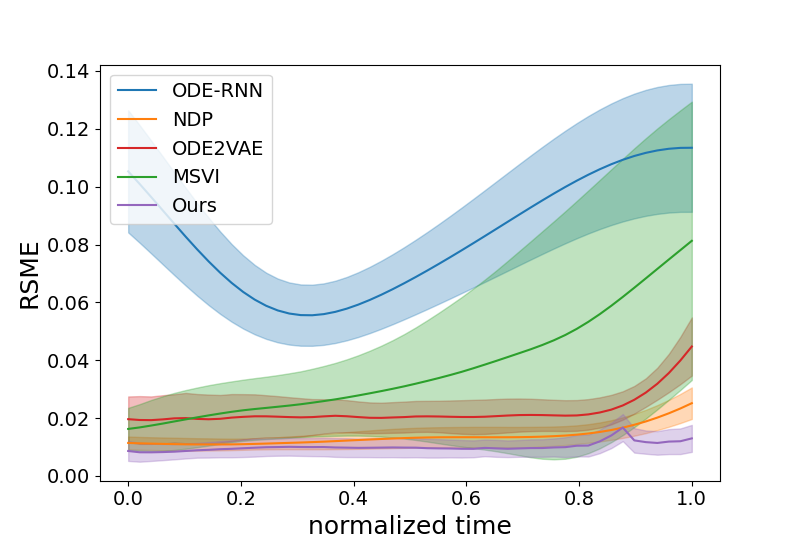}
         \caption{RMSE distribution over time}
     \end{subfigure}
     \hfill
     \begin{subfigure}[h!]{0.99\textwidth}
         \centering
         \includegraphics[width=\textwidth]{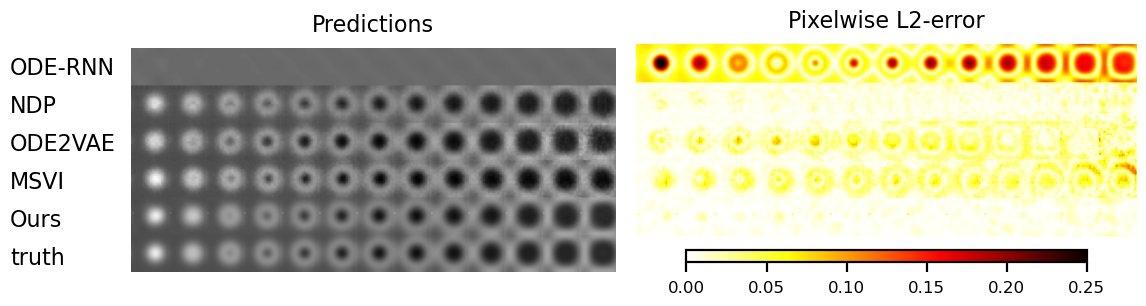}
         \caption{Example trajectory}
     \end{subfigure}
	 \caption{Test errors and exemplary test trajectories of different models for the wave equation test case.}
  \label{Fig: DINo_wave_appendix}
\end{figure}

\FloatBarrier
\newpage
\subsection{Latticed Boltzmann equations}
\begin{figure}[h!]
\begin{subfigure}[h!]{0.51\textwidth}
         \centering
         \includegraphics[width=\textwidth]{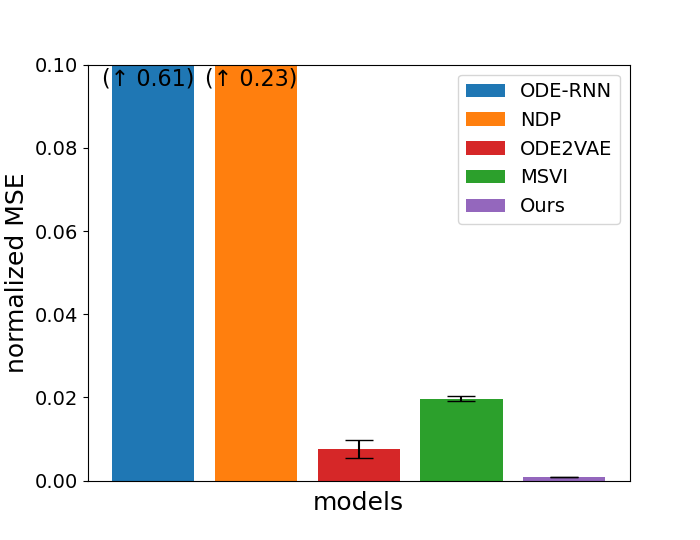}
         \caption{Normalized MSE }
     \end{subfigure}
     \hfill
     \begin{subfigure}[h!]{0.48\textwidth}
         \centering
         \includegraphics[width=\textwidth]{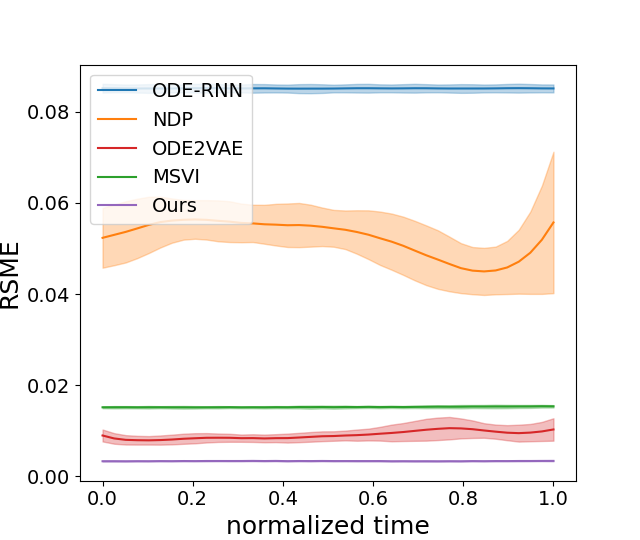}
         \caption{RMSE distribution over time}
     \end{subfigure}
     \hfill
     \begin{subfigure}[h!]{0.99\textwidth}
         \centering
         \includegraphics[width=\textwidth]{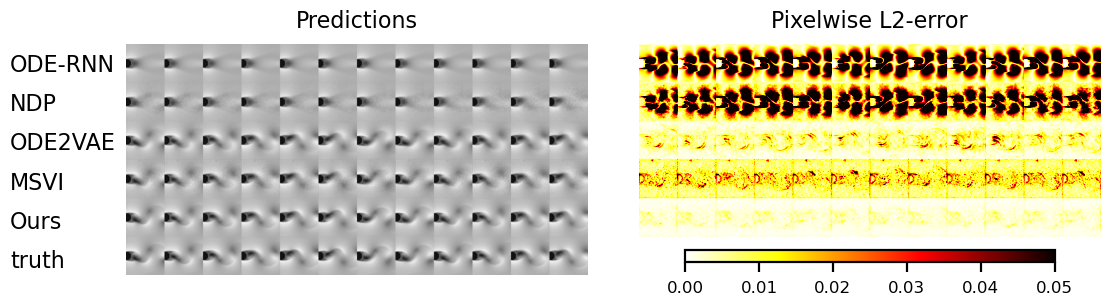}
         \caption{Example trajectory}
     \end{subfigure}
	 \caption{Test errors and exemplary test trajectories of different models for the von K\'{a}rm\'{a}n vortex street test case.}
  \label{Fig: vortex_street_appendix}
\end{figure}
\FloatBarrier

\end{document}